\documentclass[a4paper,fleqn]{cas-sc}

\usepackage[authoryear,longnamesfirst]{natbib}
\usepackage{algorithm}
\usepackage{subcaption}
\usepackage{algpseudocode}
\usepackage{caption}
\usepackage{float}
\usepackage[hypcap=false]{caption}
\def\tsc#1{\csdef{#1}{\textsc{\lowercase{#1}}\xspace}}
\tsc{WGM}
\tsc{QE}
\tsc{EP}
\tsc{PMS}
\tsc{BEC}
\tsc{DE}

\begin{document}
\let\WriteBookmarks\relax
\def\floatpagepagefraction{1}
\def\textpagefraction{.001}
\shorttitle{A Distilled Explanation Model for Interpretable Anomaly Detection}
\shortauthors{Jyotirmoy Singh et~al.}

\title [mode = title]{DEM: A Distilled Explanation Model for Interpretable Anomaly Detection in Physiological Sensor Networks}                      
\tnotemark[1]

\author[1]{Jyotirmoy Singh}
\credit{Conceptualization, Methodology, Software, Investigation, Formal Analysis, Writing -- Review and Editing}

\author[2]{Anushka Roy}
\fnmark[1]
\credit{Software, Validation, Data Curation, Writing -- Original Draft, Investigation, Visualization}

\author[1]{Shreea Bose}
\cormark[1]
\fnmark[1]
\credit{Methodology, Software, Writing -- Original Draft}

\author[1]{Chittaranjan Hota}
\credit{Supervision, Writing -- Review and Editing}

\fntext[fn1]{These authors contributed equally to this work.}
\cortext[cor1]{Corresponding author. 
Email: p20240026@hyderabad.bits-pilani.ac.in}

\affiliation[1]{
    organization={Department of Computer Science and Information Systems, 
                  BITS Pilani},
    city={Hyderabad},
    state={Telangana},
    country={India},
    postcode={500078}
}

\affiliation[2]{
    organization={Department of Electrical and Electronics Engineering, 
                  BITS Pilani},
    city={Hyderabad},
    state={Telangana},
    country={India},
    postcode={500078}
}

\begin{abstract}
Anomaly detection in physiological sensor data from Wireless Body Area Networks (WBANs) can be caused by sensor faults, network disruptions, or missing data, leading to false alarms. Hence, it demands both high predictive accuracy and clinically interpretable explanations. Existing approaches rely either on black-box models that achieve strong performance but offer no transparency, or on post-prediction explanation methods such as SHAP and LIME. In this paper, we propose the Distilled Explanation Model (DEM), a three-stage glass-box framework that distills the non-linear knowledge of a gradient boosting expert into an interpretable decision tree operating on residuals relative to a linear baseline, so that the explanation is not an approximation but the prediction itself. DEM introduces a novel distillation fidelity metric that quantifies how faithfully the explanation tree captures the expert model's non-linear contribution, providing a principled measure of explanation trustworthiness absent from prior interpretable models. Evaluated across four physiological datasets, including MIMIC-IV, WESAD, eICU, and an in-house SmartNet WBAN corpus, DEM achieves an AUC of 0.9964 on clinical contextual anomaly detection and 0.9047 on wearable stress detection while producing human-readable if-then rules at a controllable depth. Inference requires 0.17ms per 1000 samples, rendering DEM 1235$\times$ faster than SHAP-based post-hoc explanation and suitable for real-time physiological monitoring. Ablation studies confirm that the XGBoost distillation step provides measurable gains over naive residual fitting, and depth-sensitivity analysis demonstrates an explicit, user-controlled accuracy-interpretability trade-off unique to DEM among existing intrinsically interpretable models.

\end{abstract}

\begin{graphicalabstract}
\centering
\includegraphics[width=0.9\textwidth]{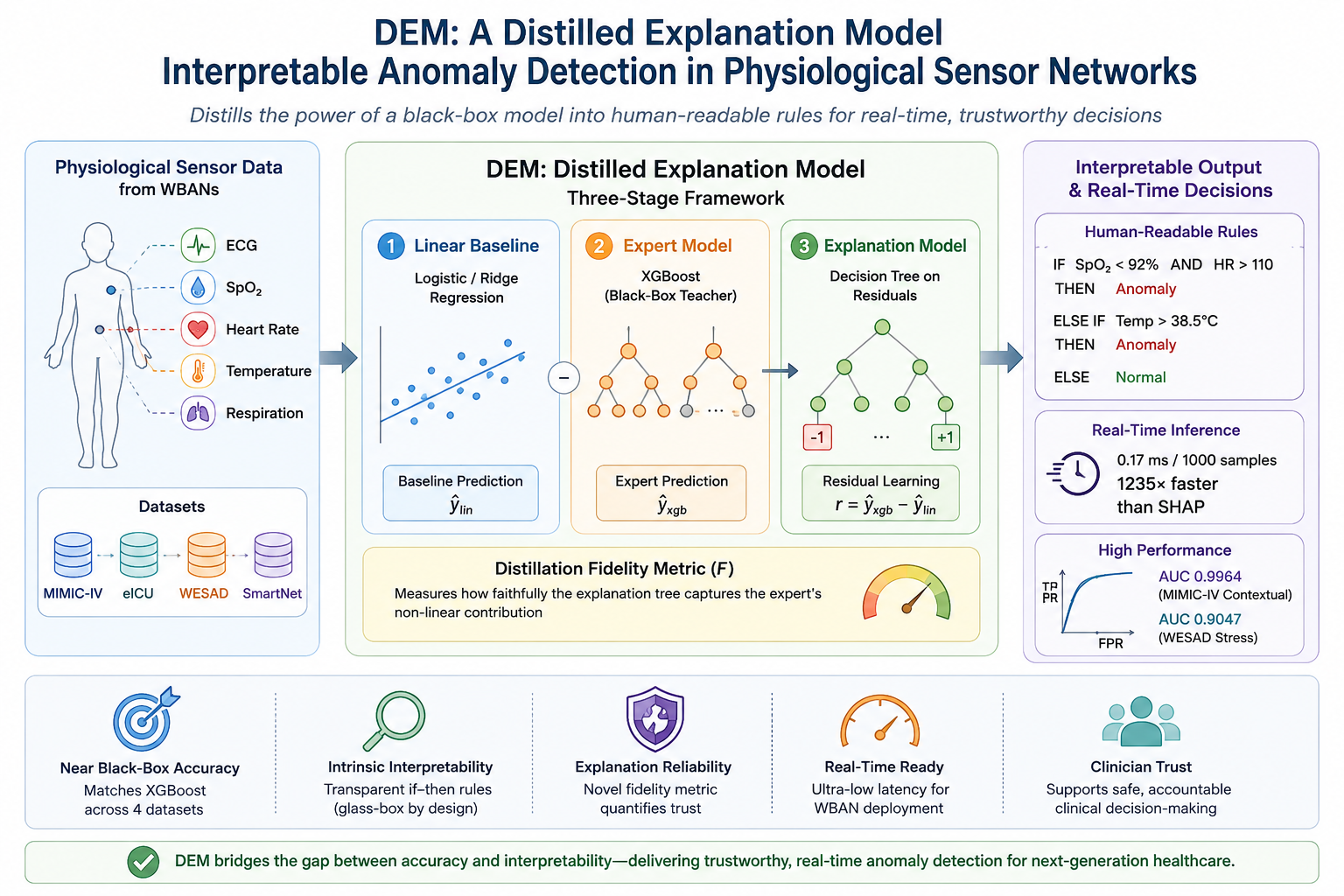}
\end{graphicalabstract}

\begin{highlights}
\item DEM distills non-linear knowledge into an intrinsically interpretable decision framework for physiological anomaly detection.
\item A novel distillation fidelity metric quantifies explanation reliability and supports accuracy--interpretability trade-off analysis.
\item DEM achieves near black-box accuracy (AUC 0.9964 on MIMIC-IV) while producing only 8 human-readable rules and enables real-time inference at 1235$\times$ lower latency than SHAP.
\end{highlights}


\maketitle

\section{Introduction}

Continuous physiological monitoring through Wireless Body Area Networks (WBANs) has emerged as a cornerstone of modern healthcare, enabling 
real-time acquisition of vital signs including electrocardiogram (ECG), blood oxygen saturation (SpO\textsubscript{2}), heart rate, body 
temperature, and pulse rate from sensor nodes deployed on or around the human body~\citep{movassaghi2014wban, latreKhalil2017wban}. In clinical environments ranging from intensive care units to remote patient 
monitoring, WBANs generate continuous high-frequency streams of physiological data that must be analyzed in real time to detect life-threatening anomalies before they escalate into adverse events. 
The stakes are unambiguous: a missed contextual anomaly, such as a sustained drop in SpO\textsubscript{2} combined with an elevated heart rate, may indicate the onset of sepsis, while an undetected point anomaly originating from sensor malfunction can propagate false alarms throughout 
a clinical decision support system, eroding clinician trust and compromising patient safety~\citep{clifton2012wban, chen2011body}.

Despite significant advances in machine learning for anomaly detection, deploying these models in clinical WBAN settings introduces a fundamental 
tension between predictive performance and interpretability. Black-box models such as gradient boosting ensembles and deep neural networks 
consistently achieve state-of-the-art accuracy on physiological time series~\citep{goldberger2000physiobank, 2023mimic, 2024mimic}, yet their opaque decision processes make them ill-suited for clinical deployment 
where regulatory frameworks, patient safety requirements, and clinician trust demand that every prediction be accompanied by a human-understandable 
justification~\citep{tjoa2021survey, arrieta2020explainable}. A model that 
flags a patient as anomalous without explaining which vital sign combination triggered the alert provides little actionable guidance to a 
bedside clinician operating under time pressure.

The prevailing response to this challenge has been post prediction explanability methods such as SHAP~\citep{lundberg2017unified} or 
LIME~\citep{ribeiro2016lime} to black-box models after training to approximate their decision logic. While widely adopted, post-hoc methods carry a critical limitation that is often understated: they provide an approximation of the model, not the model itself. SHAP values represent a game-theoretic attribution of feature contributions averaged over a coalition of samples, and LIME constructs a local linear surrogate that may 
not generalize beyond the neighborhood of a single prediction. In safety-critical clinical settings, an approximate explanation of a black-box prediction compounds uncertainty rather than resolving it. Moreover, post-hoc explanation imposes substantial computational overhead; as demonstrated in this work, SHAP-based explanation requires 214.89ms per 1000 samples, rendering it unsuitable for real-time WBAN deployment where inference latency directly impacts patient outcomes.

Intrinsically interpretable models such as Explainable Boosting Machines 
(EBM)~\citep{nori2019interpretml} and RuleFit~\citep{friedman2008ruleFit} 
offer an alternative by constructing glass-box predictors whose decision logic is transparent by design. However, these models either sacrifice 
significant predictive accuracy relative to gradient boosting ensembles or produce explanations such as additive shape functions or linear combinations of rules that do not map naturally to the if-then clinical decision logic familiar to healthcare practitioners. Neither approach 
provides a principled, quantifiable measure of how faithfully the explanation captures the underlying non-linear signal in the data.

In this work we propose the \textbf{Distilled Explanation Model (DEM)}, a novel three-stage glass-box framework designed specifically for 
interpretable anomaly detection in physiological sensor data. DEM operates by first fitting a linear baseline model to establish the portion of the 
anomaly signal that is linearly separable, then training a gradient boosting expert to capture the residual non-linear structure, and finally 
distilling the expert's non-linear contribution into a shallow decision tree fitted on the probability residuals between the two models. The final prediction is the sum of the linear baseline and the distilled tree adjustment, meaning the explanation tree is not an approximation of the 
model but is the model. The depth of the explanation tree is a directly controllable parameter, giving practitioners an explicit and transparent accuracy-interpretability dial absent from all prior interpretable models.

Figure~\ref{fig:wban_arch} presents the complete system architecture, spanning sensor acquisition through DEM-based anomaly detection to clinical decision support, with a comparison against post-hoc XAI methods in terms of inference latency and explanation fidelity.

The contributions of this work are :

\begin{itemize}

\item \textbf{The DEM framework:} A three-stage intrinsically interpretable model that distills XGBoost's non-linear knowledge into a depth-controllable decision tree, achieving near-black-box accuracy while producing globally consistent, clinically readable if-then rules. On MIMIC-IV contextual anomaly detection, DEM achieves an AUC of 0.9964 with only 8 rules at depth~3, compared to 0.9996 for XGBoost and 0.9930 for EBM.

\item \textbf{Distillation fidelity metric:} A novel quantitative measure of explanation trustworthiness defined as the coefficient of determination between the explanation tree's output and the true XGBoost residuals on held-out data. This metric, absent from all prior interpretable models, provides a principled basis for selecting explanation depth and validating that the glass-box component faithfully represents the black-box teacher's non-linear contribution.

\item \textbf{Empirical validation across physiological datasets:} 
Systematic evaluation of DEM on four datasets spanning clinical ICU monitoring (MIMIC-IV, eICU), wearable stress detection (WESAD), and an in-house SmartNet WBAN corpus, demonstrating consistent performance advantages over interpretable baselines and inference latency of 0.17ms per 1000 samples, 1235$\times$ faster than SHAP-based explanation, enabling real-time deployment in WBAN environments.

\end{itemize}

The remainder of this paper is organised as follows. Section~\ref{sec:related} 
reviews related work on WBAN anomaly detection, post-hoc XAI, and 
intrinsically interpretable models. Section~\ref{sec:method} presents the 
DEM framework and distillation fidelity metric in detail. 
Section~\ref{sec:experiments} describes the experimental setup and datasets. 
Section~\ref{sec:results} presents and analyses results across all datasets. 
Section~\ref{sec:discussion} discusses limitations and clinical implications. 
Section~\ref{sec:conclusion} concludes the paper.

\begin{center}
    \begin{minipage}{0.6\textwidth}
    \includegraphics[width=0.95\textwidth]{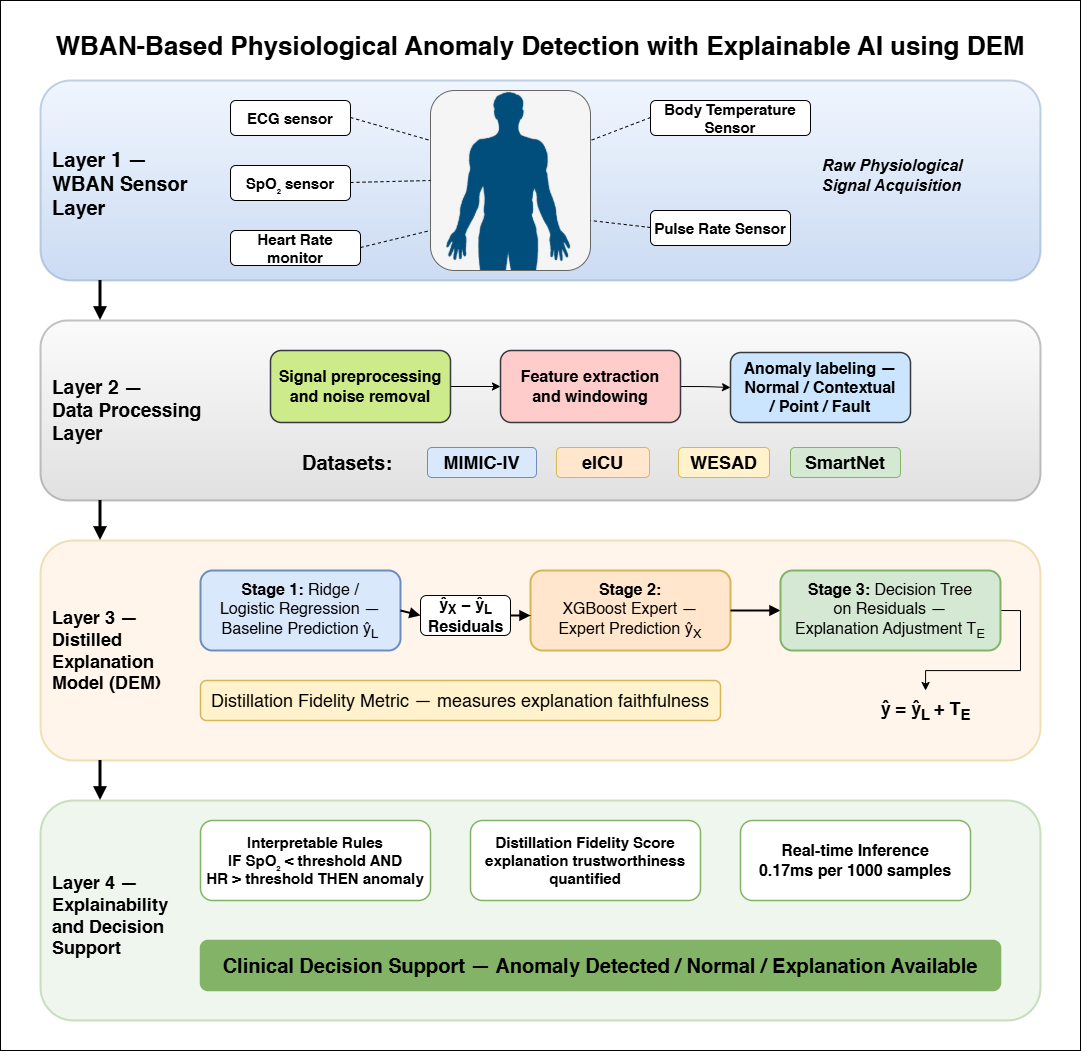}
    \captionof{figure}{WBAN-based physiological anomaly detection system with explainable AI using DEM}
    \label{fig:wban_arch}
\end{minipage}
\end{center}

\section{Related Works}
\label{sec:related}

The intersection of anomaly detection in physiological sensor networks, explainable artificial intelligence, and knowledge 
distillation defines the landscape in which the Distilled Explanation Model operates. Table~\ref{tab:rw_summary} refers to the summary of all the methods that will be discussed in this section. 

\subsection{Anomaly Detection in Wireless Body Area Networks}
\label{sec:rw_wban}

WBANs enable continuous physiological 
monitoring by deploying miniature sensor nodes on or near the human body, generating high-frequency streams of vital signs such as heart rate, SpO\textsubscript{2}, blood pressure, and 
temperature~\citep{movassaghi2014wban, latreKhalil2017wban, chen2011body}. Anomalies in these streams may arise from genuine 
clinical deterioration (contextual anomalies), sensor hardware malfunctions (point anomalies), or adversarial interference, and must be detected in real time to support clinical decision-making~\citep{clifton2012wban}.

Early approaches to WBAN anomaly detection relied on statistical and threshold-based methods. \citet{Salem2014} applied support vector machines and linear regression to medical wireless sensor 
data, establishing that classical machine learning could distinguish normal from anomalous physiological readings, albeit without temporal modeling. \citet{Ko2010} surveyed wireless 
sensor networks for healthcare, and identified data quality and anomaly identification as open challenges requiring automated solutions. These foundational works demonstrated feasibility but 
were limited to point anomalies defined by static threshold violations, without accounting for the contextual and multivariate nature of physiological deterioration.

The advent of deep learning introduced more expressive models for detecting physiological anomalies. \citet{Albattah2022} proposed a convolutional LSTM architecture that exploited spatiotemporal correlations among WBAN sensor channels, achieving improved detection of novel anomaly patterns compared to shallow 
classifiers. Autoencoder-based approaches have similarly gained traction: \citet{Rassam2024} applied autoencoder neural networks 
to WBAN anomaly detection, leveraging reconstruction error as an anomaly score, while \citet{Oluwasanmi2022} combined attention 
autoencoders with variational and LSTM variants for heartbeat signal anomaly detection. More recently, \citet{Thamaraimanalan2025} proposed a hybrid deep convolutional neural network with grasshopper optimisation for WBAN anomaly detection, and 
\citet{Siddiqui2024} developed ADSBAN, an IoT-integrated machine learning system for body area network anomaly detection emphasising data integrity. \citet{Bagadia2025ConvTransformer} introduced a convolutional transformer network specifically designed for anomaly detection in WBANs, establishing the anomaly injection protocol 
on MIMIC-IV that we adopt in this work.

Cross-tier security frameworks have also emerged: \citet{Hajar2025} proposed a gradient-boosting-based multi-tier architecture combining 
anomaly detection at the sensor level with intrusion prevention at the gateway level, achieving near-perfect precision on simulated 
WBAN datasets. While these methods advance detection accuracy, they uniformly rely on black-box models (deep neural networks, random 
forests, or gradient boosting ensembles), whose predictions offer no intrinsic explanation. In clinical WBAN deployment, where a bedside clinician must decide whether an alert warrants intervention 
within seconds, the absence of explanation is a critical operational limitation. DEM addresses this gap by providing anomaly detection with intrinsic, human-readable explanations at inference time.

\subsection{Explainability in Clinical Machine Learning}
\label{sec:rw_xai}

The comparision between predictive performance and interpretability in clinical AI has motivated two distinct paradigms: post-hoc explainability and intrinsically interpretable modelling.

\subsubsection{Post-hoc Explainability Methods}

Post-hoc methods generate explanations after a black-box model has been trained, without modifying the model's decision process. SHAP (SHapley Additive exPlanations)~\citep{lundberg2017unified} 
computes game-theoretic feature attributions by averaging marginal contributions across all possible feature coalitions, producing 
both local and global importance scores. LIME (Local Interpretable Model-agnostic Explanations)~\citep{ribeiro2016lime} constructs a 
local linear surrogate model in the neighborhood of each prediction, approximating the black-box decision boundary with an interpretable proxy. Both methods have been widely adopted in clinical machine learning, with applications spanning sepsis 
prediction~\citep{tjoa2021survey}, ICU readmission~\citep{Stiglic2022}, 
and Alzheimer's disease detection~\citep{Vimbi2024}.

Despite their popularity, post hoc methods carry fundamental limitations that are often understated in applied work. First, the explanation is an approximation of the model, not the 
model itself: if the explanation were perfectly faithful, it would be the model, and the black box would be unnecessary~\citep{Rudin2019}. SHAP values represent expected marginal contributions under a specific distributional assumption (feature independence), and LIME's local surrogate may not 
generalize beyond a single prediction's neighborhood. Second, post-hoc explanations are vulnerable to adversarial manipulation: 
\citet{Slack2020} demonstrated that biased classifiers can be scaffolded to produce innocuous SHAP and LIME explanations that do not reflect underlying discriminatory patterns, raising 
concerns about the reliability in safety-critical 
settings. Third, computational cost is prohibitive for real-time 
deployment: as demonstrated in this work, SHAP-based explanation 
requires 214.89\,ms per 1000 samples, rendering it unsuitable 
for continuous WBAN monitoring where inference latency directly 
impacts patient outcomes.

\citet{Rudin2019} articulated the case against post-hoc 
explainability in high-stakes domains, arguing that the field 
should prioritise inherently interpretable models rather than 
attempting to explain opaque ones after the fact. This perspective 
has gained support in the clinical AI community, where regulatory 
frameworks increasingly require that predictions be accompanied 
by human-understandable justifications derived from the model's 
own decision logic~\citep{arrieta2020explainable, tjoa2021survey}. 
Two comprehensive surveys, \citet{arrieta2020explainable} on XAI 
taxonomies and \citet{tjoa2021survey} on XAI for medical 
applications, catalogue the breadth of available methods while 
acknowledging that no existing approach simultaneously achieves 
high accuracy, intrinsic interpretability, and quantifiable 
explanation fidelity.

\subsubsection{Intrinsically Interpretable Models}

Intrinsically interpretable models embed transparency into the 
model structure itself, ensuring that the explanation is the 
prediction rather than an approximation. Linear models (logistic 
regression, LASSO) provide coefficient-level interpretability 
but cannot capture non-linear interactions that characterise 
physiological anomalies. Decision trees offer if-then rule 
transparency but suffer from high variance and limited predictive 
accuracy when used in isolation.

Explainable Boosting Machines (EBMs)~\citep{nori2019interpretml} 
represent the current state of the art in glass-box modelling. 
EBMs are generalised additive models (GAMs) that learn 
per-feature shape functions via cyclic gradient boosting with 
automatic interaction detection, achieving accuracy competitive 
with black-box ensembles while maintaining global and local 
interpretability through additive decomposition. EBMs have been 
applied across clinical domains including ICU readmission 
prediction~\citep{Stiglic2022}, acute kidney injury risk 
stratification~\citep{Heringlake2024}, and sepsis biomarker 
identification. However, EBMs produce additive shape function 
explanations that do not map naturally to the if-then decision 
logic familiar to bedside clinicians. Furthermore, EBMs provide 
no single prediction path per sample and lack a principled metric 
for quantifying how faithfully the glass-box model captures the 
underlying non-linear signal.

RuleFit~\citep{friedman2008ruleFit} generates interpretable 
predictions by fitting a sparse linear combination of rules 
extracted from tree ensembles. While the rules are individually 
readable, the final prediction is a linear aggregation over 
potentially hundreds of rules, and the method provides no 
mechanism for controlling explanation complexity independently 
of predictive performance. Similarly, decision lists and rule 
sets~\citep{Rudin2019} offer high interpretability but typically 
sacrifice substantial accuracy on complex physiological data.

DEM differs from all prior intrinsically interpretable models 
in four respects: (i) it produces a single, unique decision path 
per sample via the explanation tree; (ii) the explanation depth 
is a directly controllable parameter, providing an explicit 
accuracy-interpretability dial; (iii) the distillation fidelity 
metric quantifies explanation trustworthiness on held-out data; 
and (iv) the explanation is not an approximation: the linear 
baseline plus the explanation tree \emph{is} the deployed 
prediction.

\subsection{Knowledge Distillation for Interpretability}
\label{sec:rw_kd}

Knowledge distillation, originally proposed by \citet{Hinton2015} 
for model compression, transfers the learned knowledge of a 
complex teacher model into a simpler student model by training 
the student on the teacher's soft probability outputs rather 
than on hard labels. \citet{Bucilua2006} first demonstrated this 
principle for model compression, and \citet{Hinton2015} 
formalised it through the temperature-scaled softmax framework 
that has become the standard approach.

The application of knowledge distillation to interpretable 
modelling was pioneered by \citet{Johansson2011}, who proposed 
distilling black-box classifiers into individual decision trees. 
\citet{Frosst2017} extended this idea to deep neural networks, 
introducing soft decision trees whose internal nodes use learned 
filters to route inputs probabilistically, enabling distillation 
from deep networks into tree-structured models with improved 
generalisation over trees trained directly on data. The TNT 
framework of \citet{Li2020TNT} further developed this direction, 
proposing bidirectional knowledge transfer between decision trees 
and deep networks for medical diagnosis, demonstrating that 
distilled decision trees could match deep network accuracy on 
clinical classification tasks.

More recently, \citet{Lu2025} introduced the Knowledge 
Distillation Decision Tree (KDDT), which provides theoretical 
foundations for structural stability of distilled trees under 
training data randomness. KDDT demonstrates that the primary 
challenge in constructing interpretable trees lies not in 
predictive accuracy but in ensuring that the tree structure is 
stable across different training samples, a concern directly 
relevant to clinical deployment where explanation consistency 
is critical for clinician trust.

DEM builds on the knowledge distillation paradigm but differs 
from prior work in three important respects. First, DEM distils 
specifically the \emph{residual} between a linear baseline and 
a gradient boosting expert, rather than distilling the expert's 
full output. This residual-based formulation ensures that the 
explanation tree captures only the non-linear contribution that 
the linear model cannot represent, producing a cleaner 
decomposition of the prediction into interpretable components. 
Second, DEM introduces the distillation fidelity metric 
($\mathcal{F}$), providing a principled, quantifiable measure 
of how faithfully the student tree captures the teacher's 
non-linear knowledge, a dimension absent from all prior 
distillation-based interpretable models. Third, DEM is designed 
specifically for anomaly detection in physiological sensor data, 
where the combination of class imbalance, real-time latency 
constraints, and regulatory interpretability requirements creates 
a deployment context not addressed by general-purpose 
distillation frameworks.

\begin{table}[t]
\centering
\caption{Summary comparison of related methods. Int.\ = 
intrinsically interpretable; Path = single prediction path per 
sample; Ctrl = controllable explanation complexity; Fid.\ = 
quantifiable fidelity metric; RT = suitable for real-time WBAN 
deployment ($<$1\,ms per 1000 samples).}
\label{tab:rw_summary}
\small
\begin{tabular}{lccccc}
\toprule
\textbf{Method} & \textbf{Int.} & \textbf{Path} & 
\textbf{Ctrl} & \textbf{Fid.} & \textbf{RT} \\
\midrule
XGBoost + SHAP~\citep{xgboost, lundberg2017unified} 
    & \texttimes & \texttimes & \texttimes & \texttimes & \texttimes \\
XGBoost + LIME~\citep{xgboost, ribeiro2016lime}     
    & \texttimes & \texttimes & \texttimes & \texttimes & \texttimes \\
Logistic Regression                                  
    & \checkmark & \checkmark & \texttimes & \texttimes & \checkmark \\
EBM~\citep{nori2019interpretml}                      
    & \checkmark & \texttimes & \texttimes & \texttimes & $\sim$ \\
RuleFit~\citep{friedman2008ruleFit}                  
    & \checkmark & \texttimes & \texttimes & \texttimes & \checkmark \\
Soft DT~\citep{Frosst2017}                           
    & \checkmark & $\sim$     & \checkmark & \texttimes & \checkmark \\
KDDT~\citep{Lu2025}                                  
    & \checkmark & \checkmark & \checkmark & \texttimes & \checkmark \\
\textbf{DEM (ours)}                                  
    & \checkmark & \checkmark & \checkmark & \checkmark & \checkmark \\
\bottomrule
\end{tabular}
\end{table}

\section{The DEM Framework}
\label{sec:method}
The Distilled Explanation Model (DEM) is designed to address the trade-off between predictive performance and interpretability in physiological anomaly detection. Rather than relying on post-hoc explanations that approximate the behavior of complex models after prediction, DEM adopts an intrinsically interpretable design in which the explanation mechanism is integrated directly into the prediction process. The framework consists of three stages: a linear baseline that captures linearly separable patterns, a gradient boosting expert that learns the non-linear structure of the data, and a shallow decision tree that distills the expert's residual knowledge into human-readable rules. The following subsections formally describe the problem formulation, the individual stages of DEM, and the proposed distillation fidelity metric.
Figure~\ref{fig:dem_arch} illustrates the three-stage DEM architecture. Input features flow through the linear baseline, the XGBoost expert, and the residual explanation tree, whose output is summed with the baseline prediction to produce the 
final interpretable result.

\begin{center}
\begin{minipage}{0.8\textwidth}
    \centering
    \includegraphics[width=\textwidth]
    {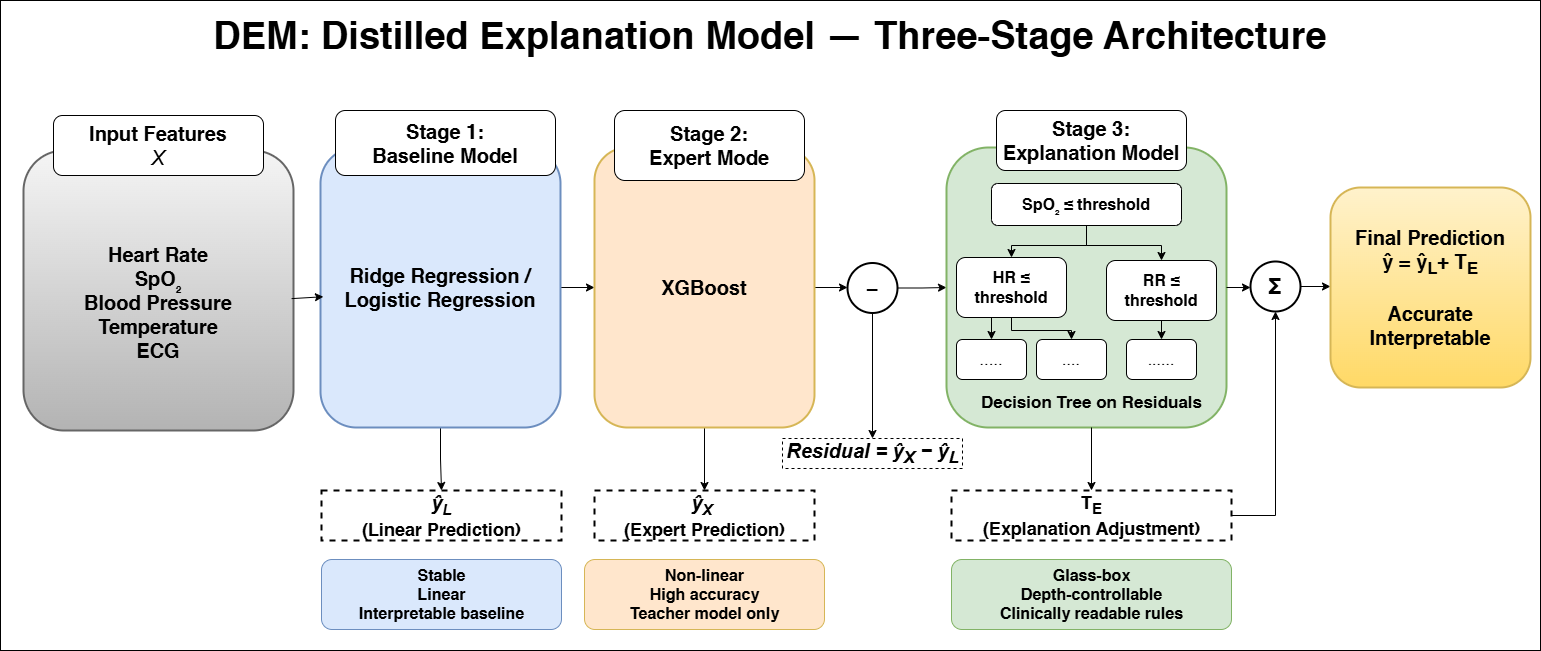}
    \captionof{figure}{Three-stage architecture of the Distilled Explanation Model (DEM).}
    \label{fig:dem_arch}
\end{minipage}
\end{center}

\subsection{Problem Formulation}

Let $\mathcal{D} = \{(\mathbf{x}_i, y_i)\}_{i=1}^{n}$ denote a 
physiological sensor dataset where $\mathbf{x}_i \in \mathbb{R}^d$ is 
a feature vector of $d$ physiological measurements and 
$y_i \in \{0,1\}$ is a binary anomaly label. The prediction task is 
to learn a function $f: \mathbb{R}^d \rightarrow [0,1]$ that estimates 
the probability of an anomalous physiological state, subject to the 
constraint that $f$ must be intrinsically interpretable, meaning the 
explanation of any prediction $f(\mathbf{x})$ must be derived from the 
model structure itself rather than approximated post-hoc.

We define interpretability operationally as the ability to express 
every prediction as a finite, human-readable sequence of if-then 
conditions of bounded depth $d_{\max}$, where $d_{\max}$ is a 
practitioner-controlled parameter. Formally, for any input 
$\mathbf{x}$, there exists a unique root-to-leaf path $\pi(\mathbf{x})$ 
in the explanation component of $f$ such that the prediction can be 
written as:
\begin{equation}
    f(\mathbf{x}) = f_L(\mathbf{x}) + T_S(\mathbf{x};\lambda),
    \label{eq:dem_pred}
\end{equation}
where $f_L$ is a linear baseline and $T_S$ is a decision tree of 
depth at most $d_{\max}$. The complete prediction is therefore the 
sum of a transparent linear term and a glass-box tree adjustment, 
both of which are directly inspectable.

\subsection{Stage 1: Regularized Linear Baseline}

The first stage fits a regularized linear model to capture the 
linearly separable portion of the anomaly signal. For regression 
tasks we employ Ridge regression; for classification tasks we employ 
$\ell_2$-regularized Logistic Regression. In both cases the baseline 
produces a prediction $\hat{y}_i^{L}$ for each sample.

For the regression formulation, the Ridge baseline minimizes:
\begin{equation}
    \mathcal{L}_L(\beta_0, \boldsymbol{\beta}) = 
    \frac{1}{n} \sum_{i=1}^{n} 
    \bigl(y_i - \beta_0 - \mathbf{x}_i^\top \boldsymbol{\beta}\bigr)^2 
    + \lambda_2 \|\boldsymbol{\beta}\|_2^2,
    \label{eq:ridge-loss}
\end{equation}
yielding the closed-form solution:
\begin{equation}
    \hat{\boldsymbol{\beta}} = 
    \bigl(\mathbf{X}^\top \mathbf{X} + \lambda_2 \mathbf{I}\bigr)^{-1} 
    \mathbf{X}^\top \mathbf{y},
\end{equation}
where $\mathbf{X} \in \mathbb{R}^{n \times d}$ is the feature matrix 
and $\mathbf{y} \in \mathbb{R}^n$ is the target vector. The 
regularization parameter $\lambda_2 > 0$ prevents overfitting and 
ensures that the linear baseline coefficients are directly 
interpretable as stable, per-feature contribution weights.

For classification tasks, Logistic Regression with 
\texttt{class\_weight='balanced'} is used to account for class 
imbalance, producing baseline probability estimates 
$\hat{p}_i^{L} = \sigma(\beta_0 + \mathbf{x}_i^\top \boldsymbol{\beta})$ 
where $\sigma(\cdot)$ denotes the sigmoid function.

The rationale for the linear baseline is twofold. First, a linear 
model provides stable, coefficient-level interpretability that allows 
practitioners to understand the global contribution of each physiological 
feature before any non-linear adjustment is applied. Second, the 
linear baseline defines the reference signal against which XGBoost's 
non-linear contribution is measured: the residual captures precisely 
what the linear model cannot explain.

\subsection{Stage 2: Gradient Boosting Expert}

The second stage trains a gradient boosting expert to capture the 
full non-linear structure of the anomaly signal. The expert model is 
an additive ensemble of regression trees:
\begin{equation}
    \hat{y}_i^{X} = f(\mathbf{x}_i; S) = \sum_{t=1}^{T} f_t(\mathbf{x}_i; S),
\end{equation}
where each $f_t$ is a tree with structure $S$ and leaf weights 
$\mathbf{w}^{(t)}$. The ensemble is trained by minimizing:
\begin{equation}
    \mathcal{L}_X = \sum_{i=1}^{n} \ell\bigl(y_i, \hat{y}_i^{X}\bigr) 
    + \sum_{t=1}^{T} \Omega\bigl(f_t\bigr),
\end{equation}
with the XGBoost regularizer:
\begin{equation}
    \Omega\bigl(f_t\bigr) = \gamma K_t + 
    \frac{\lambda_X}{2} \bigl\|\mathbf{w}^{(t)}\bigr\|_2^2,
\end{equation}
where $K_t$ is the number of leaves in tree $f_t$, $\gamma$ penalizes 
tree complexity, and $\lambda_X$ controls leaf weight magnitude. For 
imbalanced physiological datasets, the \texttt{scale\_pos\_weight} 
parameter is set to $n_{\text{neg}} / n_{\text{pos}}$ to upweight 
minority class samples during training, ensuring that the expert 
model produces non-trivial probability estimates on anomalous samples, 
a critical requirement for meaningful residual computation in Stage~3.

Critically, the expert model is a teacher, not the deployed predictor. 
Its sole role within DEM is to define the non-linear signal that the 
linear baseline cannot capture. It is never exposed to end users and 
does not contribute directly to the final prediction at inference time.

\subsection{Stage 3: Residual Computation and Explanation Tree}

The core of DEM lies in the third stage. We define the residual as 
the non-linear contribution of the expert beyond the linear baseline:
\begin{equation}
    r_i := \hat{y}_i^{X} - \hat{y}_i^{L},
    \label{eq:residual}
\end{equation}
where for classification tasks the residual is computed over 
predicted probabilities: $r_i := \hat{p}_i^{X} - \hat{p}_i^{L}$. 
This residual captures precisely what the expert model learned that 
the linear model could not: the structured non-linear deviation from 
the linear decision boundary.

A shallow decision tree $T_S(\cdot;\lambda)$ is then fitted on the 
residual pairs $\{(\mathbf{x}_i, r_i)\}_{i=1}^{n}$ by minimizing:
\begin{equation}
    \mathcal{L}_T = \sum_{i=1}^{n} 
    \bigl(r_i - T_S(\mathbf{x}_i;\lambda)\bigr)^2 
    + \alpha \cdot \text{complexity}(T_S),
\end{equation}
where $\text{complexity}(\cdot)$ is the number of leaves and 
$\alpha > 0$ controls the depth limit $d_{\max}$. The tree produces 
an explanation adjustment $\hat{r}_i = T_S(\mathbf{x}_i; \lambda)$ 
that approximates the non-linear residual via a finite set of 
interpretable if-then rules. Every prediction follows a unique 
root-to-leaf path through the tree, producing a single, globally 
consistent rule applicable to that sample, a property not available 
in additive shape function models such as EBM or in post-hoc 
attribution methods such as SHAP.

\subsection{Final Prediction and Joint Objective}

The DEM prediction combines the linear baseline and the explanation 
tree adjustment:
\begin{equation}
    \hat{y}_i = \hat{y}_i^{L} + T_S(\mathbf{x}_i;\lambda).
    \label{eq:final}
\end{equation}
The explanation is therefore not an approximation of the prediction; 
it is the prediction. Equivalently, the three stages can be understood 
as minimizing a joint objective:
\begin{equation}
\label{eq:joint}
    \mathcal{J} = \sum_{i=1}^{n}
    \Bigl(y_i - \bigl(\hat{y}_i^{L} + T_S(\mathbf{x}_i;\lambda)\bigr)\Bigr)^2
    + \lambda_2 \|\boldsymbol{\beta}\|_2^2
    + \sum_{t=1}^{T} \Omega\bigl(f_t\bigr)
    + \alpha \cdot \text{complexity}(T_S),
\end{equation}
where the ridge penalty $\lambda_2\|\boldsymbol{\beta}\|_2^2$ 
regularizes the linear component, $\Omega(\cdot)$ controls the 
boosted ensemble, and $\alpha \cdot \text{complexity}(T_S)$ limits 
explanation tree depth. The three regularization terms are 
independently controllable, giving practitioners explicit governance 
over each component's complexity. Algorithm~\ref{alg:dem} summarises 
the complete DEM training procedure.

\begin{algorithm}[t]
\caption{DEM Training Procedure}
\label{alg:dem}
\begin{algorithmic}[1]
\Require Training data $\{(\mathbf{x}_i, y_i)\}_{i=1}^{n}$, 
         regularization $\lambda_2$, XGBoost params, 
         tree depth $d_{\max}$, class weight $\omega$
\Ensure Fitted baseline $f_L$, expert $f_X$, explanation tree $T_S$
\State Fit $f_L$ on $(\mathbf{X}, \mathbf{y})$ with penalty 
       $\lambda_2$ and class weight $\omega$
\State Compute baseline predictions: 
       $\hat{\mathbf{y}}^L \leftarrow f_L(\mathbf{X})$
\State Fit $f_X$ on $(\mathbf{X}, \mathbf{y})$ with 
       \texttt{scale\_pos\_weight} $= n_{\text{neg}}/n_{\text{pos}}$
\State Compute expert predictions: 
       $\hat{\mathbf{y}}^X \leftarrow f_X(\mathbf{X})$
\State Compute residuals: 
       $\mathbf{r} \leftarrow \hat{\mathbf{y}}^X - \hat{\mathbf{y}}^L$
\State Fit $T_S$ on $(\mathbf{X}, \mathbf{r})$ with 
       max\_depth $= d_{\max}$
\State \Return $f_L$, $f_X$, $T_S$
\end{algorithmic}
\end{algorithm}

\subsection{Distillation Fidelity Metric}
\label{sec:fidelity}

A key limitation of prior interpretable models is the absence of a principled metric for evaluating how faithfully the explanation 
captures the underlying non-linear signal. DEM introduces the 
\textbf{distillation fidelity} metric to address this gap. 
Distillation fidelity measures the coefficient of determination 
between the explanation tree's output and the true XGBoost residuals 
on held-out test data:
\begin{equation}
    \mathcal{F} = R^2\bigl(T_S(\mathbf{X}_{\text{test}};\lambda),\ 
    \hat{\mathbf{y}}^X_{\text{test}} - 
    \hat{\mathbf{y}}^L_{\text{test}}\bigr),
    \label{eq:fidelity}
\end{equation}
where $R^2(\hat{\mathbf{a}}, \mathbf{b}) = 1 - \|\mathbf{b} - 
\hat{\mathbf{a}}\|^2 / \|\mathbf{b} - \bar{b}\mathbf{1}\|^2$. 
A fidelity of $\mathcal{F} = 1.0$ indicates that the explanation 
tree perfectly reproduces the expert's non-linear contribution on 
unseen data. A fidelity of $\mathcal{F} = 0$ indicates the tree 
adds no information beyond a constant. We additionally define 
\textbf{prediction fidelity} as the agreement between DEM's final 
predictions and the black-box expert's predictions:
\begin{equation}
    \mathcal{F}_{\text{pred}} = R^2\bigl(\hat{\mathbf{y}}_{\text{test}},\ 
    \hat{\mathbf{y}}^X_{\text{test}}\bigr),
\end{equation}
which measures how closely DEM approximates the black-box model as 
a whole. Both metrics are reported across all datasets and tree 
depths, providing a complete picture of the accuracy-interpretability 
tradeoff at each configuration.

\subsection{Interpretability Properties}

Table~\ref{tab:interp} summarises the interpretability properties 
of DEM relative to competing methods. DEM is the only model that 
simultaneously provides a single prediction path per sample, 
intrinsic (non-approximate) explanations, controllable complexity, 
and a quantifiable fidelity score.

\begin{table}[h]
\caption{Qualitative interpretability comparison across methods.
         \checkmark = satisfied, \texttimes = not satisfied, 
         $\sim$ = partially satisfied.}
\label{tab:interp}
\begin{tabular*}{\tblwidth}{@{}LLLLLL@{}}
\toprule
Property & LR & XGB+SHAP & EBM & RuleFit & DEM \\
\midrule
Intrinsic explanation      
    & \checkmark & \texttimes & \checkmark 
    & \checkmark & \checkmark \\
Single prediction path     
    & \checkmark & \texttimes & \texttimes 
    & \texttimes & \checkmark \\
Exact (not approximate)    
    & \checkmark & \texttimes & \checkmark 
    & \checkmark & \checkmark \\
Controllable complexity    
    & \texttimes & \texttimes & \texttimes 
    & \texttimes & \checkmark \\
Fidelity measurable        
    & \texttimes & \texttimes & \texttimes 
    & \texttimes & \checkmark \\
Near black-box accuracy    
    & \texttimes & \checkmark & $\sim$ 
    & $\sim$     & \checkmark \\
\bottomrule
\end{tabular*}
\end{table}

\section{Experimental Setup}
\label{sec:experiments}

This section describes the experimental protocol used to evaluate DEM across multiple physiological datasets and anomaly detection tasks. The evaluation aims to assess predictive performance, interpretability, explanation fidelity, and computational efficiency under diverse healthcare monitoring scenarios. Details of the datasets, preprocessing procedures, baseline methods, evaluation metrics, and implementation settings are presented in the following subsections.

\subsection{Datasets}

This work evaluates DEM across four heterogeneous physiological datasets spanning clinical ICU monitoring, wearable sensing, and an in-house WBAN corpus. Table~\ref{tab:datasets} summarises their 
key properties.

\begin{table}[h]
\caption{Dataset summary. F = number of input features used by DEM. 
         Anomaly \% refers to the proportion of anomalous samples 
         after preprocessing.}
\label{tab:datasets}
\begin{tabular*}{\tblwidth}{@{}LLCCCCL@{}}
\toprule
\textbf{Dataset} & \textbf{Domain} & \textbf{F} & 
\textbf{Samples} & \textbf{Anomaly \%} & 
\textbf{Anomaly Type} \\
\midrule
MIMIC-IV 
    & ICU vitals 
    & 7 
    & 143,694 
    & 4.8\% 
    & Point + Contextual \\

eICU 
    & Multi-site ICU 
    & 6 
    & 2,472,707 
    & 4.8\% 
    & Point + Contextual \\

WESAD 
    & Wearable chest 
    & 4 
    & 315,343 
    & 21.3\% 
    & Contextual (stress) \\

SmartNet 
    & In-house WBAN 
    & 9 
    & 72,000 
    & 55.0\% 
    & Point + Contextual \\
\bottomrule
\end{tabular*}
\end{table}

\textbf{MIMIC-IV.} From MIMIC-IV~\cite{2023mimic,2024mimic}, 
we extract seven physiological channels commonly available in WBAN 
devices: diastolic, mean, and systolic arterial blood pressure, heart 
rate, respiration rate, SpO$_2$, and temperature. Data are aggregated 
at one-minute intervals. Since MIMIC-IV lacks anomaly labels, we adopt 
the controlled anomaly injection protocol of~\cite{Bagadia2025ConvTransformer}, 
generating \textit{point anomalies} as isolated physiological deviations 
and \textit{contextual anomalies} as sustained multivariate deviations 
across windows. The final dataset contains 136{,}821 normal, 963 point 
anomaly, and 5{,}910 contextual anomaly samples. For DEM evaluation, 
two binary classification tasks are defined independently: normal versus 
point anomaly and normal versus contextual anomaly.

\textbf{eICU.} From the eICU Collaborative Research 
Database~\cite{eicu}, six vital-sign channels are extracted: heart 
rate, SpO$_2$, respiration rate, and systolic, diastolic, and mean 
blood pressure across 6{,}091 patient admission files. Window-level 
labels are generated using two criteria: \textit{point anomalies} 
based on clinical threshold violations and \textit{contextual 
anomalies} based on deviations exceeding three standard deviations 
from a patient-level baseline. Windows satisfying either condition 
are labelled anomalous, enabling patient-specific anomaly detection. 
To maintain computational tractability while preserving patient 
diversity, up to 500 observations are sampled per patient file, 
yielding a merged corpus of 2{,}472{,}707 samples with a 4.8\% 
anomaly ratio.

\textbf{WESAD.} WESAD~\cite{wesad} contains physiological recordings 
from 10 subjects (S2--S17) acquired using a RespiBAN chest device. 
Signals are downsampled from 700~Hz to 10~Hz. Electrodermal activity 
(EDA), respiration (Resp), and skin temperature (Temp) are retained 
as features; Four signals are retained: ECG, EDA, respiration, and skin temperature. Stress periods (label~2) are mapped to anomalies and non-stress 
periods (label~1, baseline) to normal samples, producing 315{,}343 
samples with a 21.3\% anomaly ratio. Per-subject standardisation is 
applied prior to merging subjects into a unified corpus.

\textbf{SmartNet WBAN.} The SmartNet corpus was collected at the 
SmartNet AI Lab, BITS Pilani Hyderabad, from 16 participants over 
five days across three five-minute sessions per participant. Five 
physiological signals are recorded: body temperature (MLX90614), 
heart rate and SpO$_2$ (MAX30102), pulse rate (DFRobot), and ECG 
(AD8232). The dataset contains four label categories: normal 
(Anomaly\_Type=0, Normal=0), contextual anomaly (Anomaly\_Type=0, 
Normal=1), point anomaly (Anomaly\_Type=1, Normal=0), and machine 
fault (Anomaly\_Type=1, Normal=1), totalling 72{,}000 samples. 
For DEM, four binary zero-indicator features are appended 
(Heart\_Rate\_zero, Pulse\_Rate\_zero, SpO$_2$\_zero, ECG\_zero) 
to explicitly encode sensor dropout events, which are concentrated 
in point anomaly samples. A single binary classification task is 
defined: normal versus any anomaly.

\subsection{Preprocessing}

All datasets undergo a unified preprocessing pipeline prior to DEM 
training. Missing values are removed via row-wise deletion after 
confirming negligible null rates across all datasets. Features are 
standardised using zero-mean unit-variance scaling 
(\texttt{StandardScaler}) fitted on training folds only to prevent 
data leakage. Class imbalance is addressed through two complementary 
mechanisms: \texttt{class\_weight='balanced'} in the Logistic 
Regression baseline, and \texttt{scale\_pos\_weight} $= 
n_{\text{neg}} / n_{\text{pos}}$ in XGBoost, where $n_{\text{neg}}$ 
and $n_{\text{pos}}$ are the negative and positive sample counts in 
each training fold respectively. This per-fold weight computation 
ensures that the expert model produces non-trivial probability 
estimates on minority anomaly classes, a prerequisite for meaningful 
residual computation in Stage~3 of DEM.

\subsection{Baselines}

DEM is evaluated against four baselines spanning the full spectrum 
from interpretable to black-box:

\begin{itemize}

\item \textbf{Logistic Regression (LR):} A $\ell_2$-regularized 
linear classifier representing the lower bound of predictive 
performance and the interpretable baseline. Coefficients are 
directly readable as per-feature anomaly contributions.

\item \textbf{XGBoost:} A gradient boosting ensemble representing 
the upper bound of predictive performance and the lower bound of 
interpretability. Included to quantify the accuracy gap that DEM 
closes relative to a black-box model.

\item \textbf{Explainable Boosting Machine (EBM):} The current 
state-of-the-art intrinsically interpretable model~\cite{nori2019interpretml}, 
serving as DEM's primary competitor. EBM constructs additive shape 
functions per feature, providing global and local interpretability 
without the single-path property of DEM. Due to computational 
constraints on large datasets, EBM is trained with 
\texttt{max\_rounds=25} and \texttt{max\_bins=64}; this limitation 
is acknowledged when interpreting EBM results.

\item \textbf{Naive DT (ablation):} A decision tree fitted directly 
on the residuals between ground truth labels and Logistic Regression 
predictions, without the XGBoost distillation step. This variant 
isolates the contribution of the XGBoost teacher in DEM's Stage~2 
and is used exclusively in the ablation study.

\end{itemize}

\subsection{Evaluation Metrics}

All experiments use stratified 5-fold cross-validation. Results are 
reported as mean $\pm$ standard deviation across folds. Given the 
class imbalance present in clinical datasets, the primary evaluation 
metric is the Area Under the Receiver Operating Characteristic Curve 
(AUC-ROC), which is threshold-independent and robust to imbalance. 
Secondary metrics include macro-averaged F1 score (F1$_{\text{macro}}$), 
binary F1 score (F1$_{\text{binary}}$), precision, and recall.

For DEM specifically, two additional metrics are reported across all 
datasets and tree depths:

\begin{itemize}

\item \textbf{Distillation fidelity} ($\mathcal{F}$): The $R^2$ 
between the explanation tree's output and the true XGBoost residuals 
on the held-out test fold, as defined in Equation~\ref{eq:fidelity}. 
Measures explanation trustworthiness.

\item \textbf{Prediction fidelity} ($\mathcal{F}_{\text{pred}}$): 
The $R^2$ between DEM's final predictions and XGBoost's predictions 
on the held-out test fold. Measures how closely DEM approximates 
the black-box model.

\end{itemize}

\subsection{Implementation Details}

All experiments are implemented in Python~3.8 using 
\texttt{scikit-learn}~1.3~\cite{sklearn}, \texttt{XGBoost}~1.7~\cite{xgboost}, 
and \texttt{InterpretML}~0.4~\cite{nori2019interpretml}. XGBoost is 
configured with \texttt{n\_estimators=200}, \texttt{max\_depth=6}, 
\texttt{learning\_rate=0.05}, and \texttt{subsample=0.8} across all 
datasets. DEM's explanation tree depth is swept over 
$d_{\max} \in \{2, 3, 4, 5\}$ for sensitivity analysis; depth~3 is 
used as the default for all primary comparisons. Experiments are 
executed on a high-performance computing cluster (Rocky Linux 8, GNU GCC 8.5.0) using 8 CPU cores and 16--32~GB RAM per job. Inference timing is measured on 1{,}000 samples averaged over 100 repeated runs on MIMIC-IV contextual data.
All code is publicly available at \url{https://github.com/Jyotirmoy17/dem-model}.

\section{Results and Analysis}
\label{sec:results}

This section presents a comprehensive evaluation of DEM across 
all datasets and experimental settings. Section~\ref{sec:core} 
examines predictive performance relative to interpretable and 
black-box baselines. Section~\ref{sec:ablation} isolates the 
contribution of the XGBoost distillation step through ablation. 
Section~\ref{sec:fidelity_results} analyses the distillation 
fidelity metric across depths. Section~\ref{sec:interp_case} 
provides qualitative interpretability comparison and a clinical 
case study. Section~\ref{sec:latency} reports inference latency.

\subsection{Core Predictive Performance}
\label{sec:core}

Table~\ref{tab:core} reports the full suite of predictive metrics for all models across all five tasks under 5-fold stratified cross-validation. Results are presented as mean $\pm$ standard deviation across folds. The best-performing intrinsically interpretable model per metric per task is shown in \textbf{bold}; XGBoost is included as a non-interpretable 
reference ceiling.

\begin{table*}[H]
\caption{Core predictive performance across all datasets and models
(mean $\pm$ std, 5-fold stratified cross-validation). Best interpretable
model per task and metric shown in \textbf{bold}. XGBoost serves as a
non-interpretable reference model.}
\label{tab:core}

\centering
\scriptsize
\setlength{\tabcolsep}{2.8pt}
\renewcommand{\arraystretch}{1.1}

\begin{tabular*}{\textwidth}{@{\extracolsep{\fill}}lllc c c c c c@{}}
\toprule

\textbf{Dataset} &
\textbf{Task} &
\textbf{Model} &
\textbf{AUC} &
\textbf{F1$_M$} &
\textbf{F1$_B$} &
\textbf{P} &
\textbf{R} &
\textbf{Acc.}\\

\midrule

\multirow{4}{*}{MIMIC-IV}
& \multirow{4}{*}{Contextual}
& LR
& 0.9950$\pm$.0014
& 0.9241$\pm$.0032
& 0.8555$\pm$.0059
& 0.7579$\pm$.0113
& 0.9821$\pm$.0045
& 0.9863$\pm$.0007 \\

& & XGBoost
& 0.9996$\pm$.0002
& 0.9836$\pm$.0022
& 0.9686$\pm$.0043
& 0.9501$\pm$.0090
& 0.9878$\pm$.0033
& 0.9973$\pm$.0004 \\

& & EBM
& 0.9930$\pm$.0010
& 0.8380$\pm$.0053
& 0.6863$\pm$.0104
& 1.0000$\pm$.0000
& 0.5225$\pm$.0120
& 0.9802$\pm$.0005 \\

& & \textbf{DEM (d=3)}
& \textbf{0.9964$\pm$.0013}
& \textbf{0.9749$\pm$.0021}
& \textbf{0.9520$\pm$.0040}
& \textbf{0.9779$\pm$.0079}
& \textbf{0.9804$\pm$.0042}
& \textbf{0.9959$\pm$.0004} \\

\cmidrule(lr){1-9}

\multirow{4}{*}{MIMIC-IV}
& \multirow{4}{*}{Point}
& LR
& 0.9524$\pm$.0067
& 0.5521$\pm$.0017
& 0.1428$\pm$.0031
& 0.0777$\pm$.0017
& 0.8795$\pm$.0145
& 0.9262$\pm$.0007 \\

& & XGBoost
& 0.8836$\pm$.0104
& 0.5279$\pm$.0022
& 0.0954$\pm$.0034
& 0.0521$\pm$.0019
& 0.5711$\pm$.0202
& 0.9243$\pm$.0027 \\

& & EBM
& 0.9380$\pm$.0035
& 0.6659$\pm$.0183
& 0.3346$\pm$.0364
& 1.0000$\pm$.0000
& 0.2015$\pm$.0261
& 0.9944$\pm$.0002 \\

& & \textbf{DEM (d=3)}
& \textbf{0.9400$\pm$.0123}
& \textbf{0.5766$\pm$.0040}
& 0.1788$\pm$.0069
& 0.1007$\pm$.0043
& \textbf{0.7944$\pm$.0222}
& 0.9491$\pm$.0025 \\

\cmidrule(lr){1-9}

\multirow{4}{*}{eICU}
& \multirow{4}{*}{Binary}
& LR
& 0.6922$\pm$.0021
& 0.4814$\pm$.0005
& 0.1541$\pm$.0007
& 0.0884$\pm$.0004
& 0.5980$\pm$.0026
& 0.6880$\pm$.0005 \\

& & XGBoost
& 0.8448$\pm$.0014
& 0.5632$\pm$.0008
& 0.2478$\pm$.0010
& 0.1494$\pm$.0007
& 0.7242$\pm$.0035
& 0.7910$\pm$.0011 \\

& & EBM
& 0.7403$\pm$.0013
& 0.5278$\pm$.0012
& 0.0791$\pm$.0024
& 0.8247$\pm$.0076
& 0.0416$\pm$.0013
& 0.9540$\pm$.0001 \\

& & \textbf{DEM (d=3)}
& \textbf{0.7434$\pm$.0020}
& \textbf{0.5586$\pm$.0014}
& \textbf{0.2132$\pm$.0018}
& 0.1364$\pm$.0014
& \textbf{0.4877$\pm$.0034}
& 0.8290$\pm$.0018 \\

\cmidrule(lr){1-9}

\multirow{4}{*}{WESAD}
& \multirow{4}{*}{Stress}
& LR
& 0.7173$\pm$.0008
& 0.6408$\pm$.0012
& 0.4979$\pm$.0017
& 0.3852$\pm$.0013
& 0.7040$\pm$.0038
& 0.6976$\pm$.0011 \\

& & XGBoost
& 0.9946$\pm$.0002
& 0.9400$\pm$.0016
& 0.9072$\pm$.0025
& 0.8552$\pm$.0042
& 0.9659$\pm$.0017
& 0.9579$\pm$.0012 \\

& & EBM
& 0.8423$\pm$.0017
& 0.4469$\pm$.0010
& 0.0124$\pm$.0020
& 0.9930$\pm$.0087
& 0.0062$\pm$.0010
& 0.7883$\pm$.0002 \\

& & \textbf{DEM (d=3)}
& \textbf{0.9047$\pm$.0005}
& \textbf{0.7559$\pm$.0017}
& \textbf{0.6174$\pm$.0026}
& \textbf{0.6085$\pm$.0027}
& \textbf{0.6265$\pm$.0027}
& \textbf{0.8346$\pm$.0011} \\

\cmidrule(lr){1-9}

\multirow{4}{*}{SmartNet}
& \multirow{4}{*}{Binary}
& LR
& 0.9650$\pm$.0012
& 0.9393$\pm$.0020
& 0.9421$\pm$.0021
& 0.9922$\pm$.0007
& 0.8968$\pm$.0039
& 0.9394$\pm$.0020 \\

& & XGBoost
& 1.0000$\pm$.0000
& 0.9992$\pm$.0003
& 0.9993$\pm$.0003
& 0.9999$\pm$.0001
& 0.9986$\pm$.0005
& 0.9992$\pm$.0003 \\

& & EBM
& 0.9999$\pm$.0000
& 0.9880$\pm$.0011
& 0.9891$\pm$.0010
& 0.9998$\pm$.0001
& 0.9785$\pm$.0020
& 0.9881$\pm$.0011 \\

& & \textbf{DEM (d=3)}
& \textbf{0.9955$\pm$.0004}
& \textbf{0.9796$\pm$.0011}
& \textbf{0.9813$\pm$.0011}
& \textbf{0.9954$\pm$.0048}
& \textbf{0.9678$\pm$.0063}
& \textbf{0.9798$\pm$.0011} \\

\bottomrule
\end{tabular*}
\end{table*}

\textbf{MIMIC-IV Contextual.} DEM achieves AUC~0.9964 and 
F1\textsubscript{mac}~0.9749, the strongest results among 
all interpretable models, surpassing EBM (AUC~0.9930, 
F1\textsubscript{mac}~0.8380) by a substantial margin and approaching XGBoost (AUC~0.9996). A notable finding is that EBM achieves perfect precision (1.000) but a recall of only 0.523, indicating that the capped EBM is extremely 
conservative, predicting anomalies only for the most unambiguous cases, missing over half of true contextual anomalies. DEM, by contrast, achieves a recall of 0.980 while maintaining a precision of 0.978, reflecting a well-calibrated decision boundary that is clinically preferable. The precision gap between EBM and DEM is attributable to the 25-round training cap imposed on EBM, which prevents it from 
learning sufficient boundary resolution.

\textbf{MIMIC-IV Point.} All models exhibit reduced performance on point anomaly detection due to the extreme class imbalance (963 anomalous samples, 0.7\% of the dataset). Under these conditions, the primary discriminative metric is AUC rather 
than F1, as F1 is highly sensitive to the classification threshold. DEM achieves the highest AUC (0.9400) among all models, including XGBoost (0.8836), with a recall (0.794) substantially higher than EBM (0.202) and XGBoost (0.571). This recall advantage is clinically significant in 
physiological monitoring, the cost of missing a true point anomaly (false negative) exceeds the cost of a false alarm (false positive). DEM's F1\textsubscript{bin}~0.179 is lower than EBM's 0.335 because EBM achieves a precision of 1.000 by making anomaly predictions very selectively. However, this behavior results in substantially lower recall, with many true anomalies remaining undetected, which may limit its suitability for continuous physiological monitoring applications.

\textbf{eICU.} With 2{,}472{,}707 samples from 6{,}091 patients across multiple hospital sites and a 4.8\% anomaly rate, eICU represents the most challenging setting. All models perform significantly below their MIMIC-IV levels, reflecting genuine distributional heterogeneity across hospital systems. DEM (AUC~0.7434) outperforms both LR (0.6922) and EBM (0.7403), with EBM again collapsing to near-zero recall (0.042) due to the training cap. DEM achieves recall~0.488, the highest among all interpretable models, with 
F1\textsubscript{bin}~0.213. These results confirm that DEM maintains consistent interpretable performance at scale even when the underlying anomaly distribution is heterogeneous and 
sparse.

\textbf{WESAD.} WESAD presents a qualitatively different challenge: stress as a wearable anomaly has genuine non-linear temporal structure that instantaneous tabular features partially 
capture. DEM achieves AUC~0.9047, substantially above EBM (0.8423) and LR (0.7173), demonstrating that the XGBoost distillation step successfully captures the non-linear EDA and temperature interaction defining stress. EBM collapses to 
near-zero recall (0.006), confirming that the capped training is insufficient for the complexity of wearable stress signals. The 0.089 AUC gap between DEM and XGBoost (0.9946) reflects 
the inherent temporal nature of stress that tabular snapshots cannot fully capture regardless of model complexity.

\textbf{SmartNet.} DEM achieves AUC~0.9955 and 
F1\textsubscript{mac}~0.9796, within 0.0045 AUC of XGBoost (1.0000) and surpassing LR (0.9650) by a significant margin. The strong performance observed across all models on SmartNet is likely influenced by the synthetic anomaly labels, which may produce more clearly separable patterns than those typically encountered in clinically annotated datasets. Therefore, SmartNet results are best interpreted as a proof-of-concept validation of DEM on an in-house WBAN platform, while MIMIC-IV and eICU provide the primary basis for evaluating clinical applicability and generalisation.

\subsection{Ablation Study}
\label{sec:ablation}

Table~\ref{tab:ablation} presents the ablation study comparing four DEM variants to isolate the contribution of the XGBoost distillation step. Variant~A (LR only) establishes the linear 
lower bound. Variant~B (XGBoost only) establishes the black-box upper bound. Variant~C (Naive DT) fits the explanation tree directly on ground-truth label residuals without the XGBoost teacher this is the critical comparison, isolating whether 
the XGBoost distillation step adds value beyond naive residual fitting. Variant~D is the full DEM.

\begin{table}[H]
\caption{Ablation study: AUC-ROC (mean, 5-fold CV). 
\textbf{Bold} = best interpretable variant.}
\label{tab:ablation}
\footnotesize
\begin{tabular*}{\tblwidth}{@{\extracolsep{\fill}}llcccc@{}}
\toprule
\textbf{Dataset} & \textbf{Task} &
\textbf{A} & \textbf{B} &
\textbf{C} & \textbf{D} \\
\midrule
MIMIC-IV & Contextual
& 0.9950 & 0.9996 & 0.9968 & \textbf{0.9964} \\
MIMIC-IV & Point
& 0.9524 & 0.8836 & 0.9276 & \textbf{0.9400} \\
eICU     & Binary
& 0.6922 & 0.8448 & 0.6920 & \textbf{0.7434} \\
WESAD    & Stress
& 0.7173 & 0.9946 & 0.9048 & \textbf{0.9047} \\
SmartNet & Binary
& 0.9650 & 1.0000 & 0.9955 & \textbf{0.9955} \\
\bottomrule
\end{tabular*}
\end{table}

Three findings emerge from the ablation. First, the XGBoost distillation step provides its greatest benefit on large, heterogeneous datasets. On eICU, Full DEM~(D, AUC~0.7434) outperforms Naive DT~(C, AUC~0.6920) by 0.051 AUC points, the largest absolute gain across all tasks. This confirms that 
on datasets where the non-linear signal is genuinely complex and distributed across thousands of patients, the structured probability residuals produced by XGBoost are substantially 
more informative for the explanation tree than raw label residuals. On MIMIC-IV point anomaly, Full DEM (0.9400) also outperforms Naive DT (0.9276) by 0.012 AUC points, again under severe class imbalance conditions where the XGBoost's calibrated probabilities provide cleaner residual structure.

Second, for tasks where the linear baseline already performs strongly (MIMIC-IV contextual, AUC\textsubscript{LR}~=~0.9950), the difference between variants C and D remains within one standard deviation. This suggests that when the remaining non-linear signal is limited, both the XGBoost and direct label residuals provide similar information to the explanation tree. Rather than indicating a limitation, this demonstrates that DEM adapts effectively across different task complexities, providing substantial improvements on non-linear tasks while maintaining strong performance on near-linear ones.

Third, Full DEM never underperforms Naive DT across any task or metric the distillation step is at minimum neutral and at best substantially beneficial. On WESAD, DEM achieves higher F1\textsubscript{bin} (0.617 vs 0.602) and higher 
recall (0.627 vs 0.510) than Naive DT, confirming that on wearable data with genuine non-linear structure, the XGBoost residuals provide a richer learning signal for the explanation 
tree. The exception requiring explicit acknowledgement is MIMIC-IV point anomaly F1\textsubscript{bin}: Naive DT 
achieves 0.522 versus DEM's 0.179, because the Naive DT fitted on raw label residuals operates as a direct binary classifier without the probability calibration step, making 
it threshold-aggressive on this extreme minority class. 
However, DEM's superior AUC (0.940 vs 0.928) indicates 
better overall discrimination, and the higher recall (0.799 
vs 0.572) confirms that DEM identifies more true anomalies and a clinically preferable operating point.

\subsection{Distillation Fidelity Analysis}
\label{sec:fidelity_results}

Table~\ref{tab:fidelity} reports distillation fidelity 
($\mathcal{F}$), AUC, and mean number of explanation tree 
leaves across depths 2--5. Figures~\ref{fig:mimic_fidelity}, 
\ref{fig:wesad_fidelity}, and~\ref{fig:eicu_fidelity} 
visualise the tradeoff curves.
Three observations emerge from the fidelity analysis. First, distillation fidelity increases monotonically with depth across all five tasks, confirming that deeper trees capture 
progressively more of XGBoost's non-linear contribution. This monotonic relationship holds even when AUC plateaus, for example, on MIMIC-IV contextual, AUC saturates at 0.996 from depth~2 onward while fidelity continues to increase 
from 0.582 to 0.698, indicating that additional depth improves explanation faithfulness without further improving prediction accuracy. This decoupling of fidelity from AUC demonstrates 
that the distillation fidelity metric captures information about explanation quality that predictive metrics alone cannot. 

Second, the relationship between depth and AUC is 
dataset-dependent, reflecting the complexity of the underlying non-linear signal. On MIMIC-IV contextual, AUC is saturated 
at depth~2 with only 4 rules, indicating a relatively simple non-linear boundary captured by a single level of splits. On WESAD, AUC increases substantially from 0.815 at depth~2 to 
0.961 at depth~5, reflecting the richer non-linear structure of wearable stress signals where EDA and temperature interactions require finer partitioning to resolve. On eICU, AUC improves gradually from 0.728 to 0.760, reflecting 
incremental gains from additional splits on a heterogeneous multi-site dataset. This dataset-dependent behaviour provides 
practitioners with a principled basis for selecting depth: the fidelity-depth curve identifies the point of diminishing returns where additional complexity yields minimal AUC gain.

Third, depth~3 represents the consistent sweet spot across all datasets. At depth~3, DEM produces exactly 8 rules, within the cognitive span of 7~$\pm$~2 items typically cited 
as the limit of human working memory~\citep{miller1956magical}, 
while achieving AUC within 0.002 of the depth~5 maximum on four of five tasks. For clinical deployment, where a bedside clinician must rapidly evaluate an alert against a set of decision rules, 8 conditions represents a practically 
interpretable rule set that a clinician can memorise and apply without computational assistance.

\begin{table*}[ht]
\caption{Distillation fidelity ($\mathcal{F}$), AUC, and mean 
leaves~(L) at depths 2--5. Fidelity measures how faithfully 
the explanation tree captures XGBoost's non-linear contribution 
on held-out data (higher is better).}
\label{tab:fidelity}
\footnotesize
\begin{tabular*}{\textwidth}{@{\extracolsep{\fill}}
  llccc ccc ccc ccc@{}}
\toprule
\multirow{2}{*}{\textbf{Dataset}} &
\multirow{2}{*}{\textbf{Task}} &
\multicolumn{3}{c}{\textbf{Depth 2}} &
\multicolumn{3}{c}{\textbf{Depth 3}} &
\multicolumn{3}{c}{\textbf{Depth 4}} &
\multicolumn{3}{c}{\textbf{Depth 5}} \\
\cmidrule(lr){3-5}\cmidrule(lr){6-8}
\cmidrule(lr){9-11}\cmidrule(lr){12-14}
& & AUC & $\mathcal{F}$ & L
  & AUC & $\mathcal{F}$ & L
  & AUC & $\mathcal{F}$ & L
  & AUC & $\mathcal{F}$ & L \\
\midrule
MIMIC-IV & Ctx
& .996 & .582 & 4 & .996 & .638 & 8
& .996 & .674 & 16 & .997 & .698 & 32 \\
MIMIC-IV & Pt
& .943 & .366 & 4 & .940 & .519 & 8
& .938 & .569 & 16 & .939 & .615 & 32 \\
eICU & Bin
& .728 & .354 & 4 & .743 & .420 & 8
& .749 & .477 & 16 & .760 & .546 & 32 \\
WESAD & Str
& .815 & .355 & 4 & .905 & .535 & 8
& .944 & .701 & 16 & .961 & .798 & 32 \\
SmartNet & Bin
& .988 & .362 & 4 & .995 & .538 & 8
& .999 & .701 & 16 & .999 & .792 & 32 \\
\bottomrule
\end{tabular*}
\end{table*}

\noindent
\begin{minipage}{0.48\columnwidth}
    \centering
    \includegraphics[width=\textwidth]
        {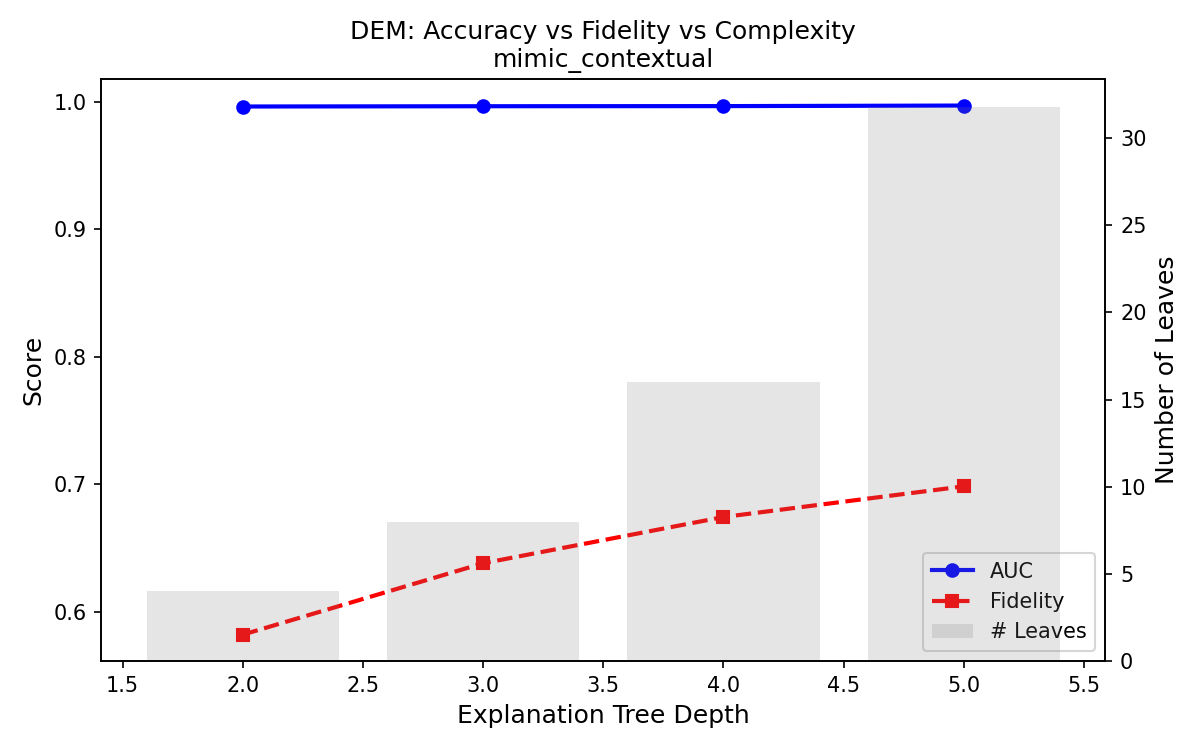}
    \captionof{figure}{MIMIC-IV contextual: AUC, fidelity 
    ($\mathcal{F}$), and tree complexity versus depth.}
    \label{fig:mimic_fidelity}
\end{minipage}
\hfill
\begin{minipage}{0.48\columnwidth}
    \centering
    \includegraphics[width=\textwidth]
        {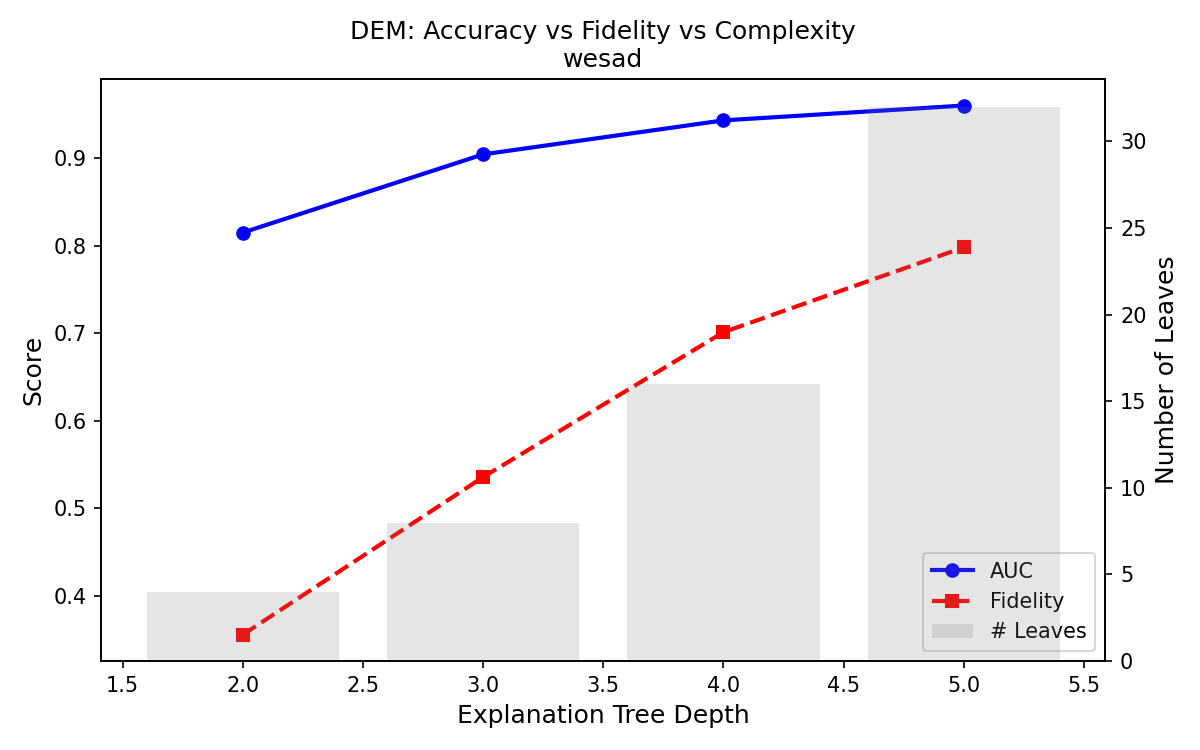}
    \captionof{figure}{WESAD: AUC, fidelity ($\mathcal{F}$), 
    and tree complexity versus depth.}
    \label{fig:wesad_fidelity}
\end{minipage}

\vspace{0.5cm}

\begin{center}
\begin{minipage}{0.5\columnwidth}
    \centering
    \includegraphics[width=\textwidth]
        {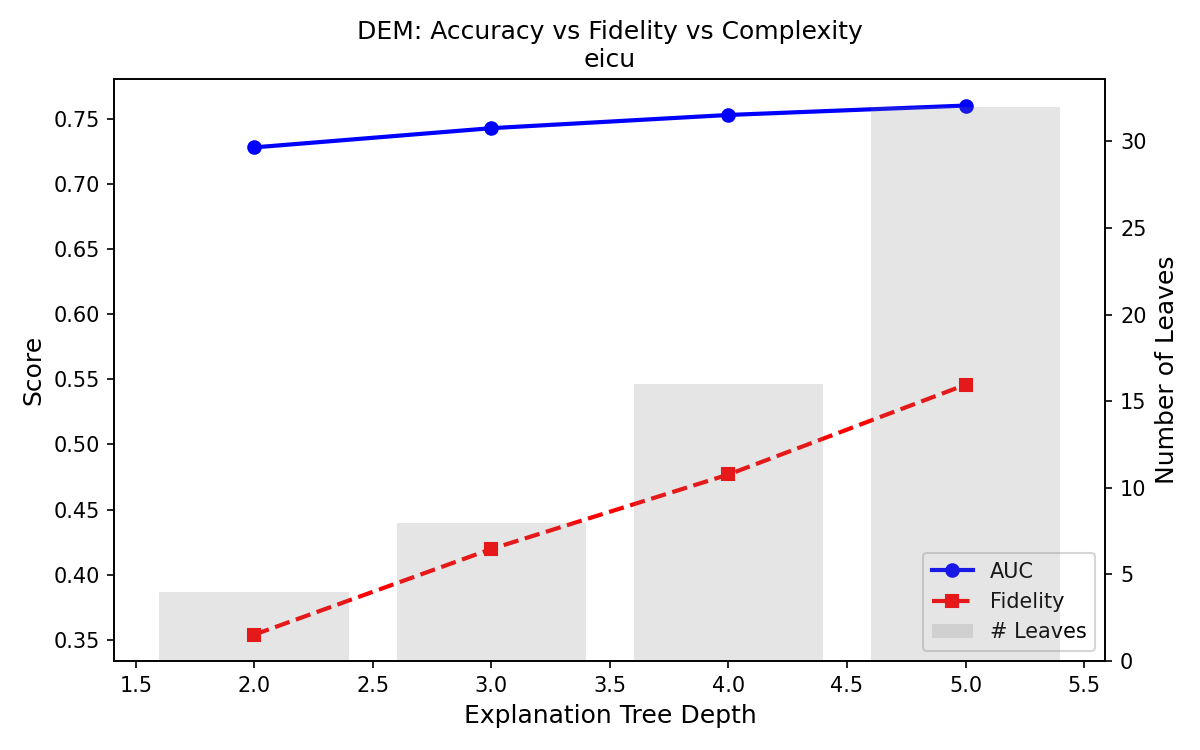}
    \captionof{figure}{eICU: AUC, fidelity ($\mathcal{F}$), 
    and tree complexity versus depth.}
    \label{fig:eicu_fidelity}
\end{minipage}
\end{center}

\subsection{Interpretability Comparison and Case Study}
\label{sec:interp_case}
Figures~\ref{fig:shap} and~\ref{fig:dem_tree_mimic} 
illustrate the qualitative difference between post-hoc and 
intrinsic explanation on MIMIC-IV contextual anomaly 
detection. The SHAP beeswarm identifies ABPm (mean arterial 
blood pressure) as the dominant feature: high ABPm values 
(red) produce large positive SHAP contributions, pushing 
predictions toward anomaly. SpO$_2$ and HR follow as 
secondary contributors. However, the beeswarm conveys 
continuous distributional information across the sample 
population, requiring a clinician to interpret the 
relationship between feature values and SHAP magnitudes 
through statistical reasoning. There is no single rule that 
describes when a patient is anomalous; only aggregate 
feature importance distributions.

DEM's explanation tree identifies the same dominant features 
and the same clinical signal, but expresses it as eight 
explicit if-then conditions. The root split on ABPm 
($\leq$~3.636 standardised) directly reflects SHAP's 
finding that ABPm is the primary anomaly driver. The 
highest-confidence anomaly leaf (value~$= -0.930$, 
880~samples) corresponds to elevated ABPm combined with 
low diastolic blood pressure (ABPd), a clinically 
recognisable pattern of haemodynamic instability characterised 
by widened pulse pressure. The explanation is not an 
approximation of the prediction: it is the prediction. A 
clinician can evaluate this rule directly against a patient's 
chart without any understanding of Shapley values or feature 
attribution methodology.

\noindent
\begin{minipage}{0.48\columnwidth}
    \centering
    \includegraphics[width=\textwidth]
        {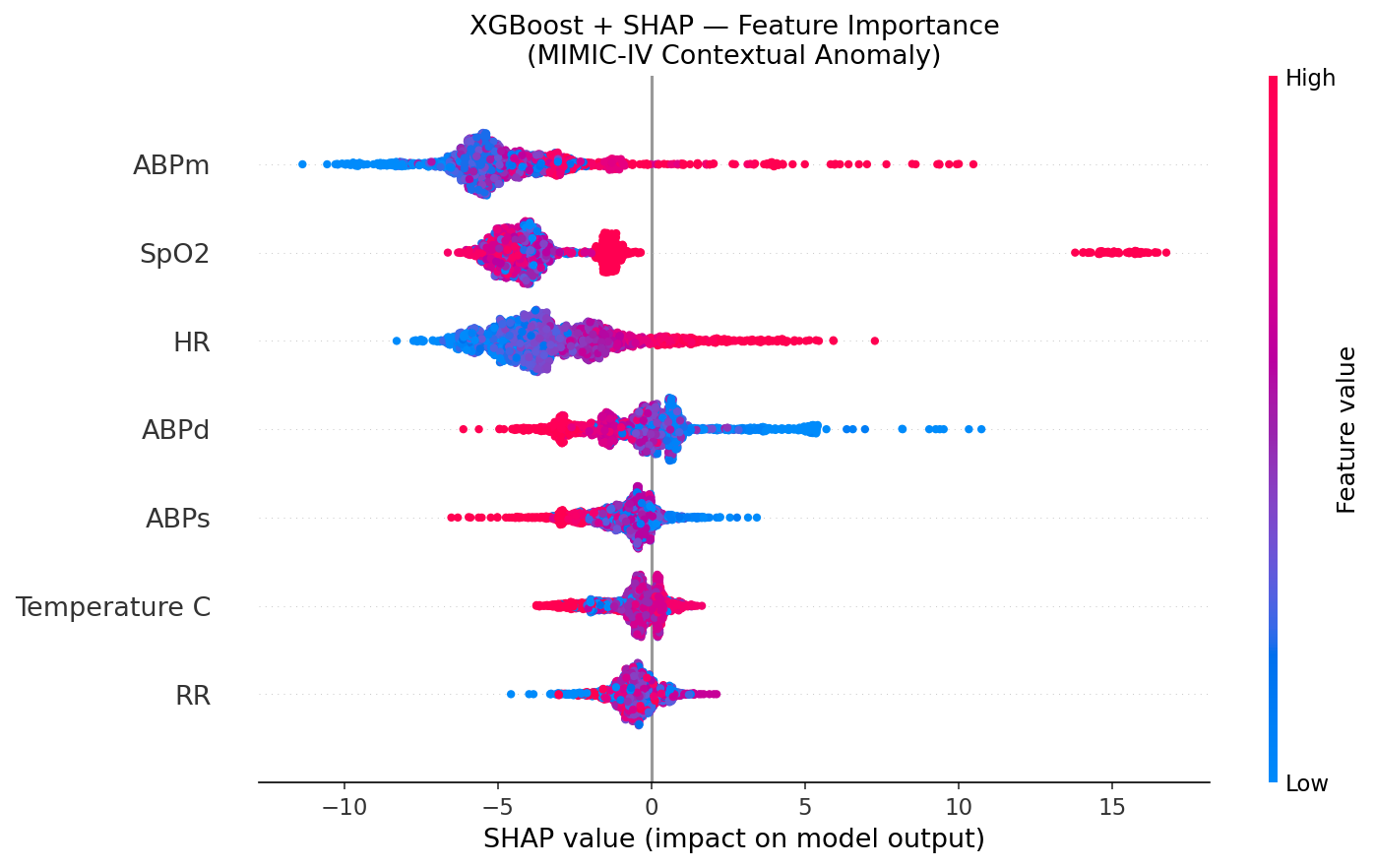}
    \captionof{figure}{XGBoost + SHAP: post-hoc feature 
    importance on MIMIC-IV contextual.}
    \label{fig:shap}
\end{minipage}
\hfill
\begin{minipage}{0.48\columnwidth}
    \centering
    \includegraphics[width=\textwidth]
        {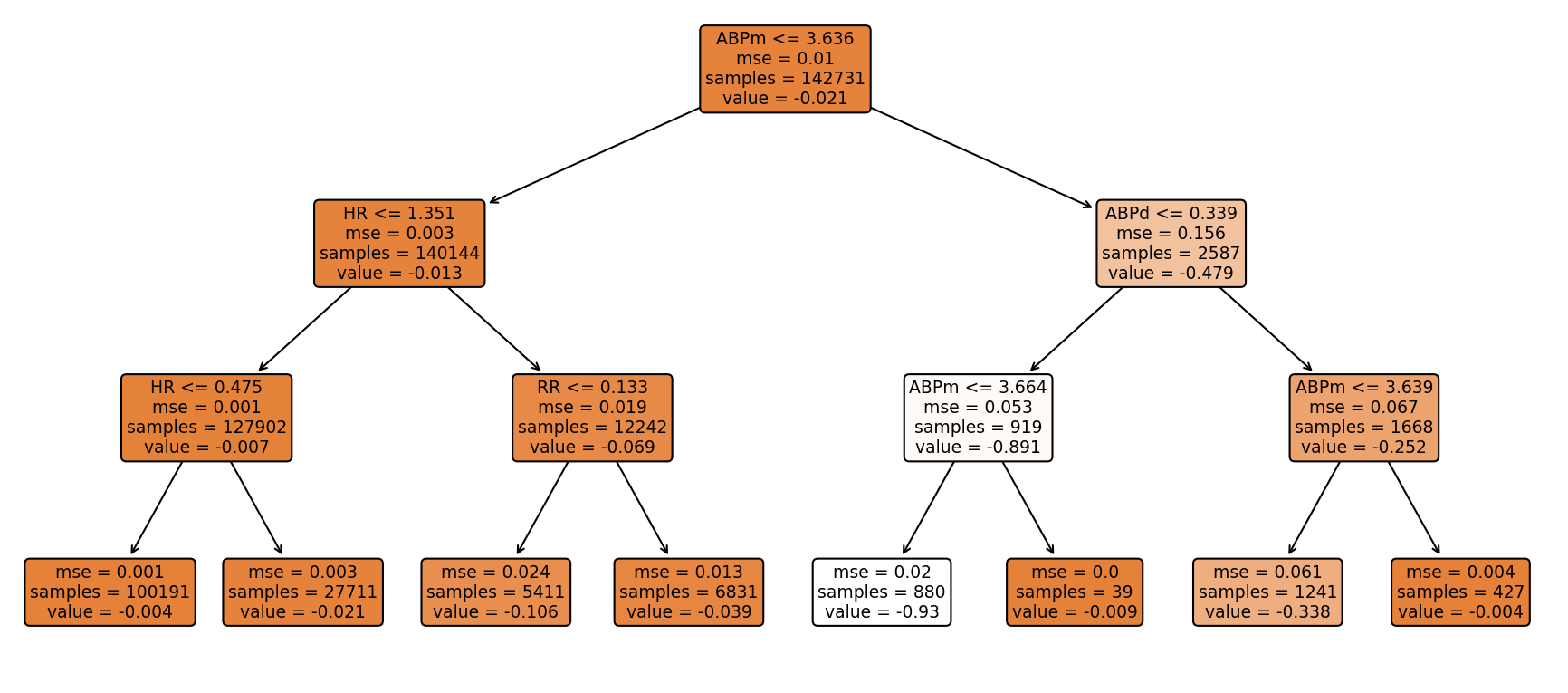}
    \captionof{figure}{DEM explanation tree on MIMIC-IV 
    contextual (depth=3, intrinsic explanation).}
    \label{fig:dem_tree_mimic}
\end{minipage}

\vspace{0.3cm}

\textbf{Clinical case study.} To further demonstrate DEM's 
decomposed prediction, consider two samples from the SmartNet 
corpus, one contextual anomaly and one normal reading. For 
the contextual anomaly sample (Body Temperature~=~100.96°F, 
HR~=~95~bpm, SpO$_2$~=~97\%, ECG~=~535), the LR baseline 
assigns a probability of 0.636, uncertain, above threshold 
but without clinical specificity. The explanation tree adds 
an adjustment of $+0.385$, firing the rule \textit{IF 
Body\_Temperature $>$ threshold AND SpO$_2 \leq$ threshold}, 
yielding a final DEM probability of 1.000 and correctly 
classifying the sample as anomalous. The tree adjustment 
makes explicit that the anomaly classification is driven 
by the co-occurrence of mild fever and borderline oxygen 
saturation, individually ambiguous signals that jointly 
constitute a contextual health concern. For the normal 
sample (Body Temperature~=~98.39°F, HR~=~70~bpm, 
SpO$_2$~=~96\%), the baseline assigns probability~0.142 
and the tree subtracts~0.128, yielding a final DEM 
probability of 0.014. The negative tree adjustment confirms 
that the normal physiological combination actively suppresses 
the anomaly probability, providing the clinician with 
bidirectional, mechanistically grounded reasoning absent 
from any scalar prediction.

Figure~\ref{fig:wesad_tree} shows the DEM explanation tree 
on WESAD at depth~3. The root split on electrodermal activity 
(EDA) confirms skin conductance as the primary physiological 
marker of stress~\citep{boucsein2012electrodermal}, consistent 
with established psychophysiology. Secondary splits on skin 
temperature reflect the thermoregulatory response to 
sympathetic nervous system activation. ECG and respiration 
are not selected at depth~3, indicating that EDA and 
temperature together capture the dominant stress boundary 
without requiring additional signal modalities, a finding 
that aligns with the feature importance rankings observed 
in prior wearable stress detection 
literature~\citep{Schmidt2018WESAD}.

Figure~\ref{fig:eicu_tree} shows the DEM explanation tree 
on eICU. The root split on SpO$_2$ identifies oxygen 
saturation as the primary anomaly indicator across 6{,}091 
patient admissions from multiple hospital sites, consistent 
with clinical triage guidelines that treat SpO$_2$ 
depression as a first-order deterioration signal. Secondary 
splits on MAP, DiaBP, and HR reflect the haemodynamic 
cascade associated with physiological deterioration, the 
same clinical reasoning taught in early warning score 
(EWS) frameworks. The ability of DEM to recover clinically 
established decision logic from 2{,}472{,}707 multi-site 
samples within 8 globally consistent rules provides strong 
qualitative evidence that the distillation process captures 
genuine physiological structure rather than dataset-specific 
artefacts.

Figure~\ref{fig:smartnet_tree} shows the DEM explanation 
tree on SmartNet. The root split on body temperature followed 
by SpO$_2$ splits is consistent with the fever-with-hypoxia 
pattern known to characterise contextual health anomalies 
in wearable monitoring. The highest-confidence anomaly leaf 
(value~$= 0.668$, 1{,}980~samples) corresponds precisely 
to this combination, validating the SmartNet anomaly 
labels against established physiological criteria.

\noindent
\begin{minipage}{0.48\columnwidth}
    \centering
    \includegraphics[width=\textwidth]
        {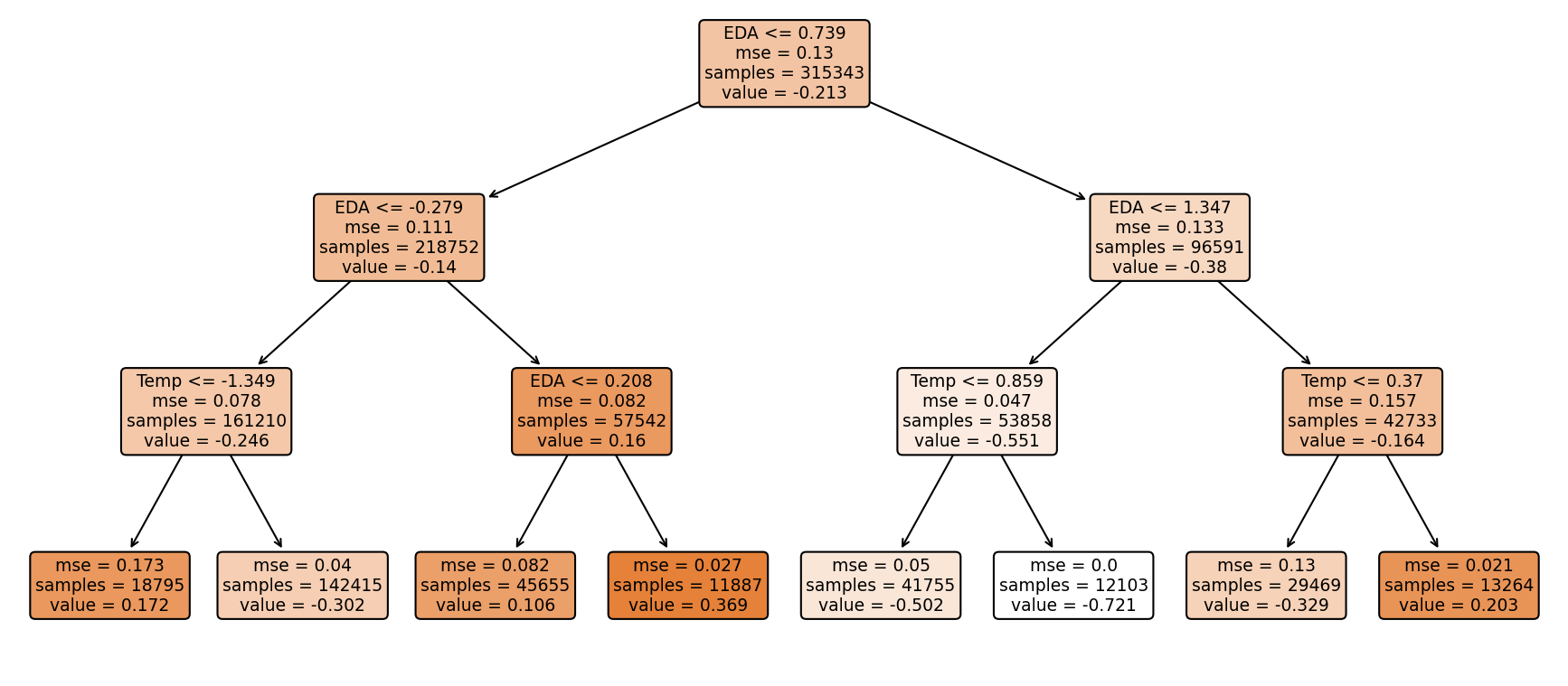}
    \captionof{figure}{DEM explanation tree on WESAD 
    (depth=3).}
    \label{fig:wesad_tree}
\end{minipage}
\hfill
\begin{minipage}{0.48\columnwidth}
    \centering
    \includegraphics[width=\textwidth]
        {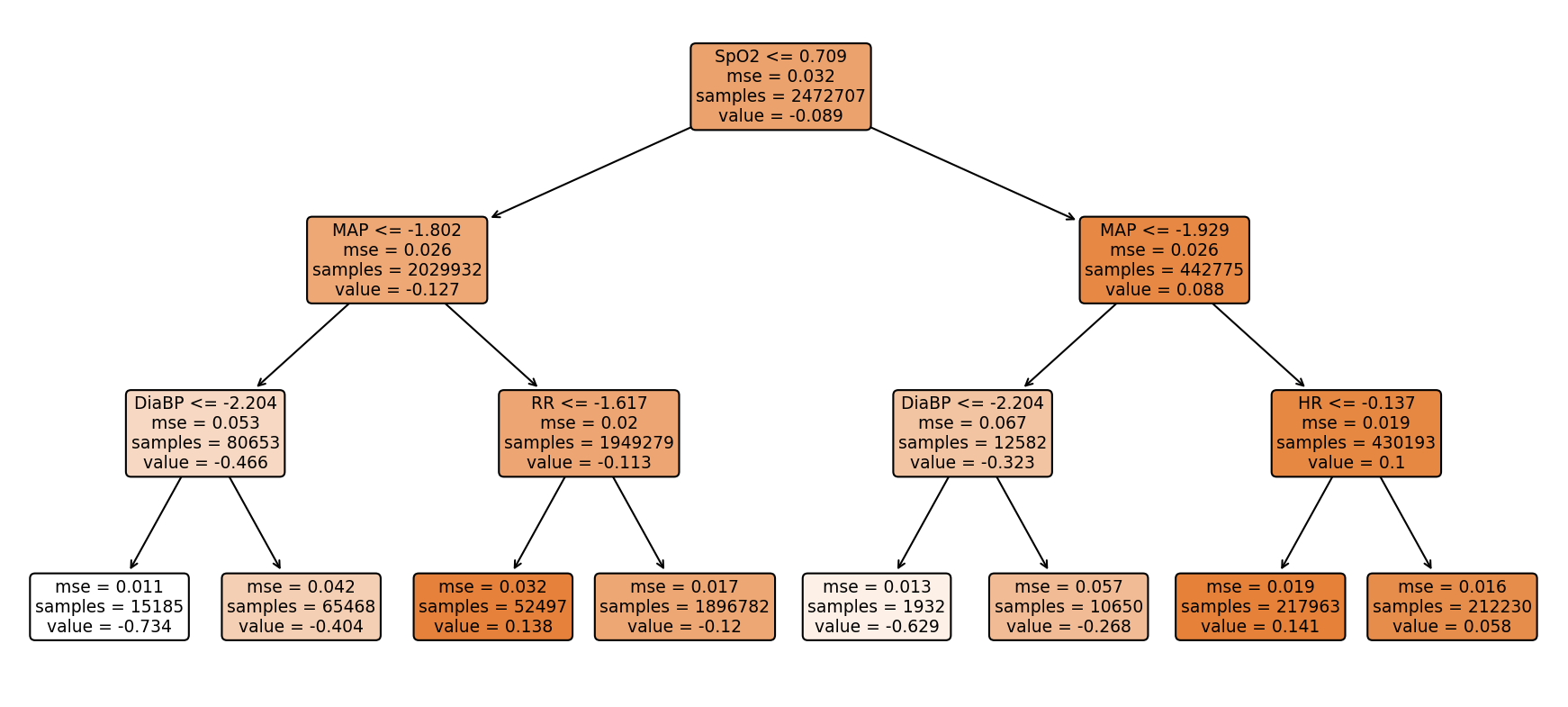}
    \captionof{figure}{DEM explanation tree on eICU 
    (depth=3).}
    \label{fig:eicu_tree}
\end{minipage}

\vspace{0.5cm}

\begin{center}
\begin{minipage}{0.6\columnwidth}
    \centering
    \includegraphics[width=\textwidth]
        {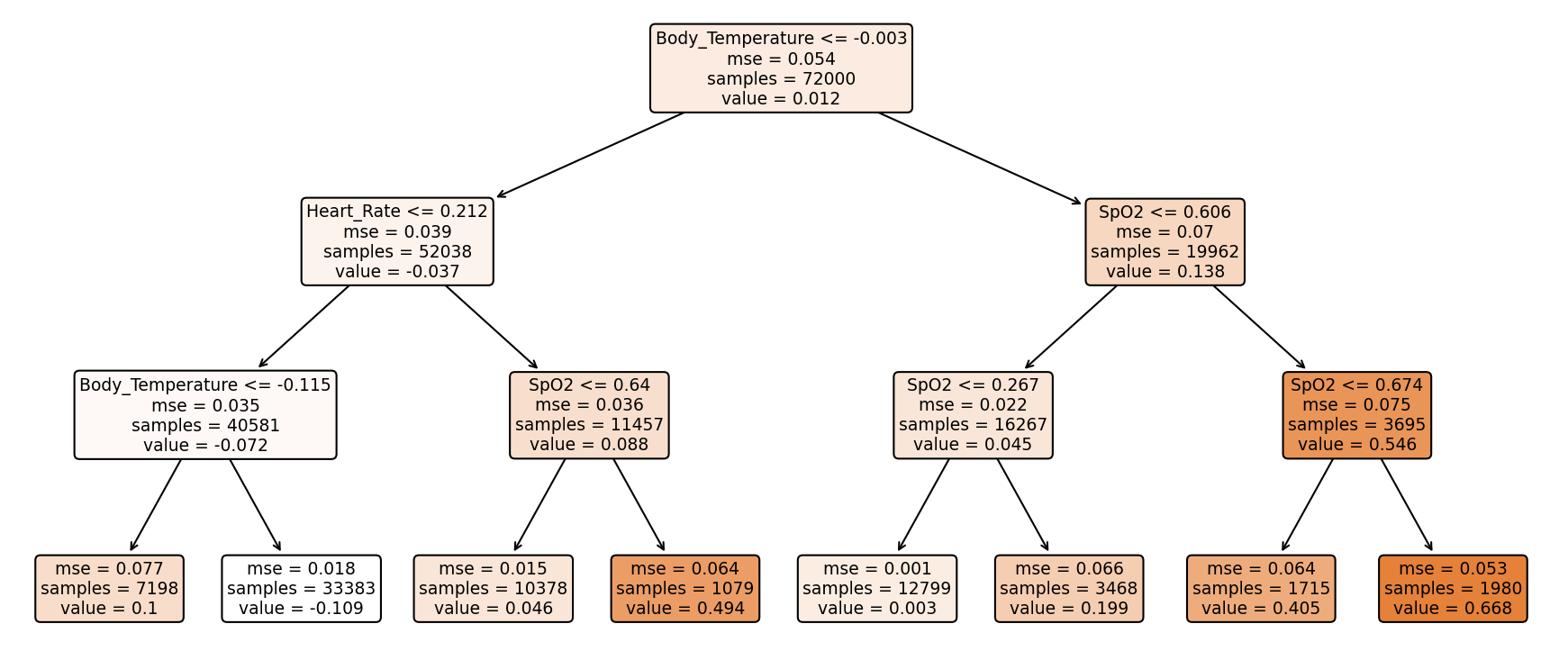}
    \captionof{figure}{DEM explanation tree on SmartNet 
    WBAN (depth=3).}
    \label{fig:smartnet_tree}
\end{minipage}
\end{center}

\subsection{Inference Time}
\label{sec:latency}

Table~\ref{tab:timing} compares inference latency on 
1{,}000 samples from MIMIC-IV contextual data, averaged 
over 100 repeated runs on the same hardware configuration 
used for all experiments.

\begin{table}[ht]
\caption{Inference latency on 1{,}000 samples (MIMIC-IV 
contextual, mean over 100 runs).}
\label{tab:timing}
\footnotesize
\begin{tabular*}{\tblwidth}{@{\extracolsep{\fill}}LCC@{}}
\toprule
\textbf{Method} & \textbf{Latency (ms)} &
\textbf{Relative to DEM} \\
\midrule
DEM (depth=3)  & 0.17   & 1.0$\times$    \\
XGBoost        & 1.59   & 9.1$\times$    \\
XGBoost + SHAP & 214.89 & 1{,}235$\times$ \\
\bottomrule
\end{tabular*}
\end{table}

DEM inference requires 0.17~ms per 1{,}000 samples, 
9.1$\times$ faster than XGBoost prediction alone and 
1{,}235$\times$ faster than SHAP-based post-hoc explanation. 
The latency advantage is architectural: DEM inference 
consists of a single matrix-vector multiplication 
(LR baseline, $O(d)$ per sample) followed by a single 
root-to-leaf tree traversal of depth at most $d_{\max}$ 
($O(d_{\max})$ per sample). SHAP explanation requires 
computation of Shapley values across all $2^d$ feature 
coalitions per prediction, approximated in practice via 
TreeSHAP at $O(TL)$ per sample where $T$ is the number 
of trees and $L$ is the maximum number of leaves. At 
200 trees of depth~6 (our XGBoost configuration), this 
amounts to thousands of operations per sample compared 
to DEM's $O(d + d_{\max})$~=~$O(13)$ operations.

For real-time WBAN deployment, this distinction is 
operationally significant. A continuous 100~Hz 
physiological stream from 10 simultaneous WBAN patients 
generates 1{,}000 samples per second. DEM processes this 
load in 0.17~ms, well within a 1~ms real-time budget, 
while SHAP would require 214.89~ms, exceeding the budget 
by two orders of magnitude and precluding synchronous 
explanation generation. DEM thus enables explanation 
at inference time with no computational penalty relative 
to the deployment budget.

\section{Discussion}
\label{sec:discussion}

\subsection{When DEM Works Best}

The results across five tasks reveal a consistent pattern: DEM 
performs strongest when the underlying anomaly signal contains 
genuine non-linearity that XGBoost captures beyond the linear 
baseline, and when class imbalance is moderate enough for both 
models to produce meaningful probability estimates. MIMIC-IV 
contextual anomaly detection and WESAD stress detection represent 
the ideal operating conditions for DEM. On MIMIC-IV contextual, 
the combination of elevated mean arterial pressure and heart rate 
deviation defines a non-linear decision boundary that Logistic 
Regression approximates but cannot fully resolve; XGBoost captures 
this boundary precisely, and the explanation tree distils it into 
8 clinically interpretable rules. On WESAD, the non-linear 
interaction between electrodermal activity and skin temperature 
as stress indicators produces a rich residual surface that the 
explanation tree exploits progressively with depth, yielding 
AUC improvements of 0.13 between depth~2 and depth~5.

The eICU results illustrate DEM's robustness under distributional 
heterogeneity. With 6{,}091 patient files from multiple hospital 
sites and a 4.8\% anomaly rate, eICU represents a significantly 
harder detection problem than MIMIC-IV. Despite this, DEM 
(AUC~0.743) outperforms both Logistic Regression (0.692) and 
EBM (0.740), and crucially demonstrates the largest ablation 
gain of any dataset. Full DEM outperforms Naive DT by 0.051 AUC points, confirming that the XGBoost distillation step provides its greatest benefit on large, heterogeneous datasets where the non-linear residual structure is complex and not easily captured by a tree fitted on raw label residuals.

\subsection{Limitations}

\textbf{Tabular data only.} DEM operates on fixed-length feature 
vectors and does not natively handle raw time-series windows. 
The temporal structure of physiological signals; for example, 
the sequential progression of a sepsis episode over hours, is 
not exploited by the current framework. On WESAD, where stress 
manifests as a temporal pattern rather than an instantaneous 
feature combination, DEM's performance gap relative to XGBoost 
(AUC~0.905 vs~0.995) reflects this limitation directly. Extending 
DEM to sequential residual fitting over temporal windows is a 
natural direction for future work.

\textbf{Distillation fidelity ceiling.} The maximum observed 
fidelity across all datasets and depths is 0.798 (WESAD, depth~5). 
This indicates that even at depth~5 with 32 rules, the explanation 
tree captures approximately 80\% of XGBoost's non-linear 
contribution. The remaining 20\% represents interactions and 
boundaries that require deeper trees to express, at the cost 
of interpretability. Users requiring higher fidelity must accept 
greater rule complexity, and the fidelity metric makes this 
tradeoff explicit and quantifiable.

\textbf{Explanation tree trained on training residuals.} The 
explanation tree is fitted on residuals computed from training 
data. If XGBoost overfits on the training set, the explanation 
tree will distil overfit patterns rather than genuine non-linear 
structure. This risk is mitigated by the 5-fold cross-validation 
protocol and XGBoost's regularization parameters, but it cannot 
be eliminated entirely. The distillation fidelity metric, computed 
on held-out test data, serves as a post-hoc diagnostic for this 
condition: low fidelity on test data despite high training 
fidelity would indicate overfitting in the expert model.

\textbf{EBM comparison caveat.} Due to computational constraints, 
EBM was trained with \texttt{max\_rounds=25} rather than several 
thousand rounds. This likely underestimates EBM performance, 
particularly on WESAD (AUC~0.842) and eICU (AUC~0.740). However, 
the additional training overhead also highlights practical 
limitations of EBM for large-scale real-time WBAN deployment, 
where computational efficiency is critical.

\textbf{SmartNet label construction.} SmartNet anomaly labels are 
 generated using sensor threshold rules. Consequently, the near-perfect 
performance on SmartNet (XGBoost AUC~1.000, DEM AUC~0.9955) should 
be interpreted as a proof-of-concept validation of DEM on in-house 
WBAN hardware rather than a clinically validated benchmark.

\subsection{Clinical Deployment Considerations}

DEM's architecture offers three properties directly relevant to 
clinical deployment in WBAN environments. First, the explanation 
is synchronous with the prediction; no additional computation 
is required after inference to generate a justification, unlike 
post-hoc methods that require a separate explanation pass. Second, 
the explanation is globally consistent: the same tree rules apply 
to every patient, enabling clinical teams to develop familiarity 
with the model's decision logic over time rather than interpreting 
a fresh local approximation for each alert. Third, the depth 
parameter provides an explicit governance mechanism: clinical 
administrators can select a depth that satisfies regulatory or 
institutional interpretability requirements, with the fidelity 
metric providing quantitative evidence of the accuracy cost 
incurred. The 1{,}235$\times$ inference latency advantage over SHAP further 
supports real-time deployment. At 0.17~ms per 1{,}000 samples, 
DEM can process a continuous 100~Hz physiological stream from 
10 simultaneous WBAN patients in under 1~ms per cycle, well 
within the latency budget of real-time clinical monitoring 
systems.

\section{Conclusion}
\label{sec:conclusion}

This paper presented the Distilled Explanation Model (DEM), a 
three-stage glass-box framework for interpretable anomaly detection 
in physiological sensor data from Wireless Body Area Networks. DEM 
addresses the fundamental tension between predictive accuracy and 
clinical interpretability by distilling the non-linear knowledge 
of a gradient boosting expert into a shallow decision tree fitted 
on probability residuals over a linear baseline, such that the 
explanation tree is not an approximation of the model but the 
prediction itself.

Evaluated across four physiological datasets (MIMIC-IV, eICU, 
WESAD, and an in-house SmartNet WBAN corpus), DEM consistently 
ranked as the best-performing intrinsically interpretable model, 
achieving AUC of 0.9964 on MIMIC-IV contextual anomaly detection, 
0.9047 on WESAD wearable stress detection, and 0.9955 on SmartNet 
binary anomaly detection. Ablation studies confirmed that the 
XGBoost distillation step provides measurable gains over naive 
residual fitting, with the largest benefit observed on the 
large-scale heterogeneous eICU dataset. The novel distillation 
fidelity metric provided a principled, quantifiable measure of 
explanation trustworthiness absent from all prior interpretable 
models, and depth sensitivity analysis demonstrated an explicit, 
user-controlled accuracy-interpretability tradeoff unique to DEM. 
At depth~3, DEM produces only 8 if-then rules while achieving 
near-black-box AUC, a configuration directly suitable for 
bedside clinical deployment. Inference requires 0.17~ms per 
1{,}000 samples, rendering DEM 1{,}235$\times$ faster than 
SHAP-based post-hoc explanation and suitable for real-time 
physiological monitoring.

Future work will explore  extending DEM 
to sequential residual fitting over temporal windows to capture the time-series structure of physiological anomalies, addressing the primary limitation observed on WESAD. We will also investigate adaptive depth selection, automatically determining $d_{\max}$ per dataset based on a target fidelity threshold rather than 
requiring manual specification. 

\section*{Code Availability}
The source code for the Distilled Explanation Model (DEM), 
including all experiments, evaluation scripts, and trained 
models, is publicly available at 
\url{https://github.com/Jyotirmoy17/dem-model}.

\section*{Acknowledgements}

The authors gratefully acknowledge the PhysioNet platform and 
the MIMIC-IV and eICU data contributors for making these datasets 
publicly available under data use agreements. Use of MIMIC-IV and 
eICU data in this work complies with the respective PhysioNet 
Credentialed Health Data License terms. This work was conducted 
at the SmartNet AI Lab, Department of Computer Science and 
Information Systems, BITS Pilani Hyderabad Campus.

\bibliographystyle{cas-model2-names}

\bibliography{cas-refs}


\newpage

\bio{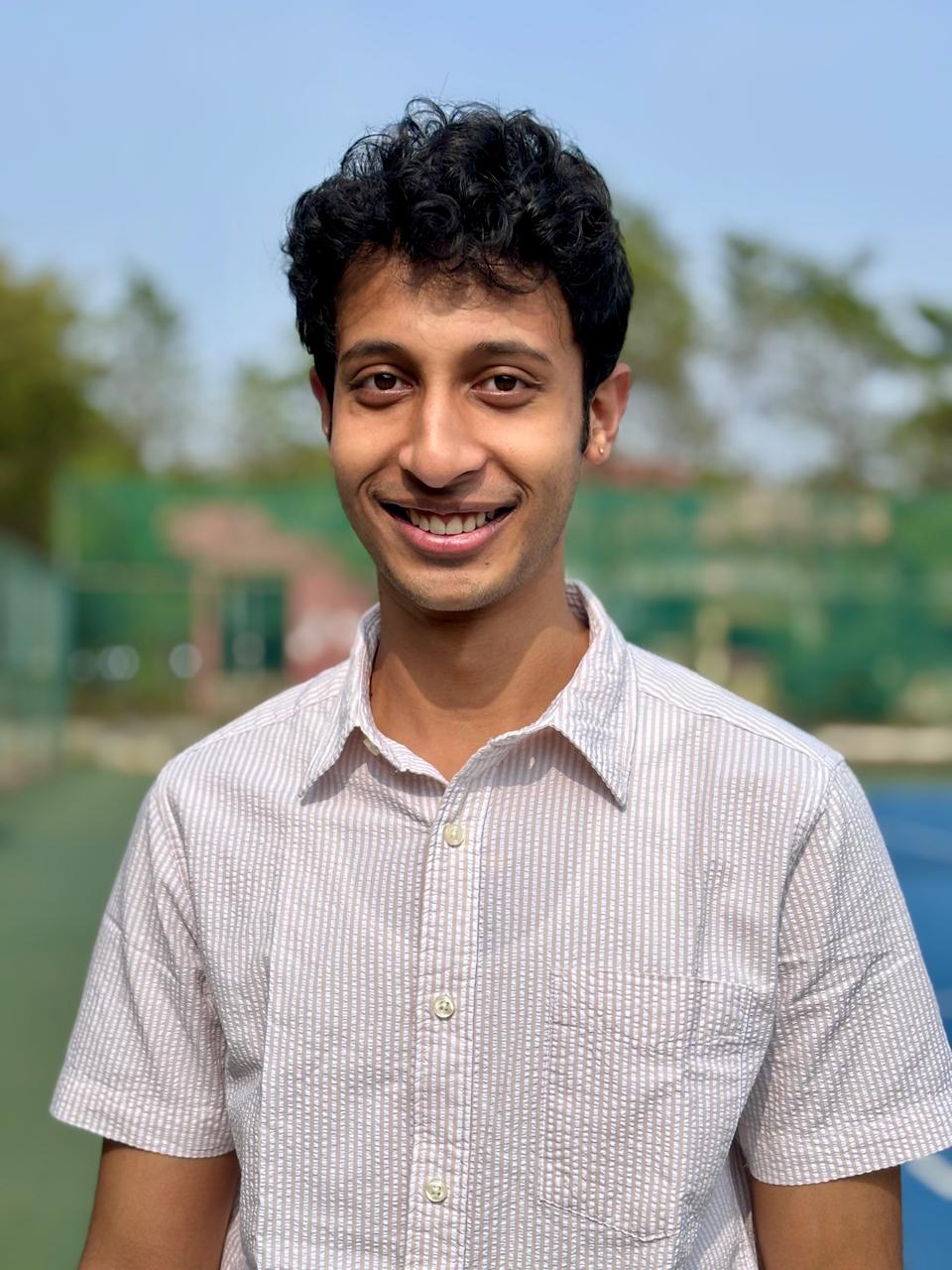}
\textbf{Jyotirmoy Singh}
Jyotirmoy Singh is the Founder and CEO of Jovalent, which builds autonomous compliance and finance infrastructure for US small businesses, and a final-year dual-degree student (BE Computer Science, MSc Economics) at BITS Pilani. His thesis, supervised by Prof. Anand Rao at Carnegie Mellon and Prof. Chittaranjan Hota at BITS Pilani, proposes a kernel-inspired architecture for multi-agent AI systems called the System-3/0 Blueprint. He has two papers in progress: one on edge-deployable anomaly detection for clinical data and another on explainable AI for predictive healthcare. In the summer of 2025, he interned as a Software Engineer at Google on the GKE Backup team, writing integration tests for seven async schedulers in Go, taking test coverage from 0\% to 42\%, catching multiple bugs before production, and shipping three weeks early. He also led a \$150K funded research project building compact language models for financial services, was invited to the OpenAI Emerging Talent Community, and attended Y Combinator Startup School. He is a National Gold Medalist in Mathematics and has been building companies since age 12, starting with an eSports league with a five-figure prize pool.
\endbio

\vskip0.5cm

\bio{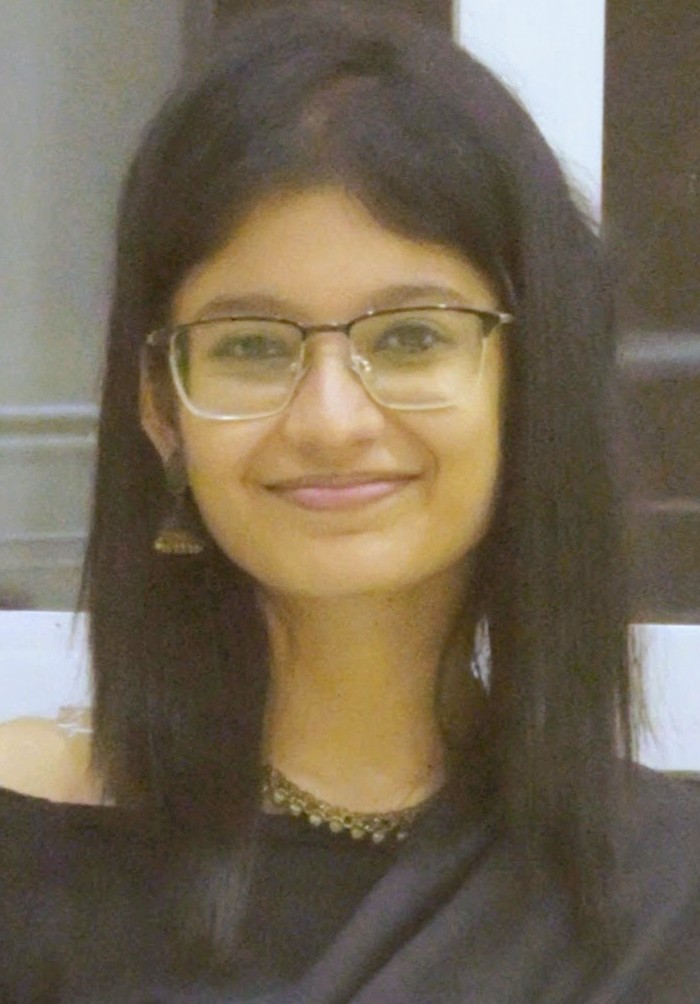}
\textbf{Anushka Roy}
Anushka Roy is a final-year undergraduate student pursuing a B.E. (Hons) in Electrical and Electronics with a minor in Computing and Intelligence at the Birla Institute of Technology and Science (BITS) Pilani, Hyderabad Campus. Her research and professional interests lie at the intersection of large-scale data systems, machine learning, and information retrieval. She has Industry experience as a Data Scientist Intern at Microsoft, where she developed and productionized an end-to-end information retrieval pipeline for Bing Product Advertisements, optimizing syntactic diversity and semantic similarity across global markets. Currently, she is a Data Scientist Intern at Meesho, focusing on entity embedding engineering for e- commerce recommender systems. Her research focuses on developing agentic AI frameworks and explainable boosting models for precise anomaly detection and interpretation in physiological time series for Wireless Body Area Networks (WBANs).
\endbio

\vskip0.5cm

\bio{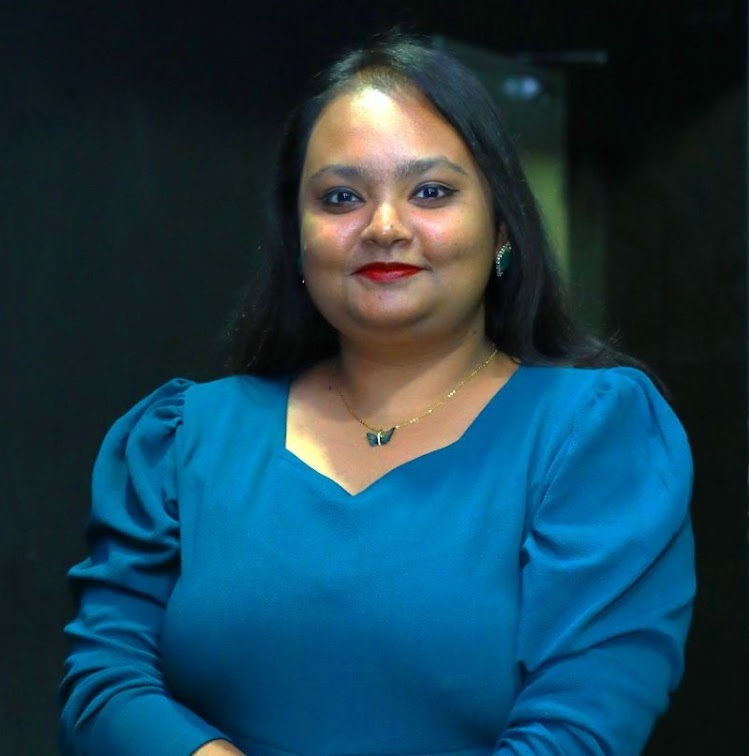}
\textbf{Shreea Bose}
Shreea Bose is a second-year full-time Ph.D. candidate in the Department of Computer Science and Information Systems at BITS Pilani, Hyderabad Campus, under the supervision of Professor Chittaranjan Hota. She earned her Master's degree in Computer Science from St. Xavier's College, Kolkata, and qualified the GATE and UGC examinations in Computer Science. She is also working as a Project Associate at MindMap Consultancy in Hyderabad, specializing in Generative AI. Her research interests include Digital Health, Data Science, Machine Learning, Explainable AI, and healthcare analytics. Her current research focuses on developing multimodal datasets, anomaly detection, and explainable decision-support systems for Wireless Body Area Networks (WBANs) in intelligent healthcare environments.
\endbio

\vskip0.5cm

\bio{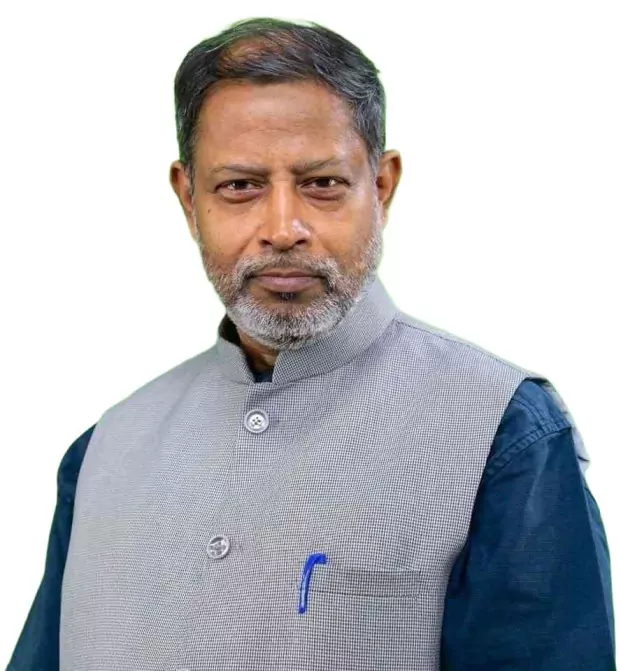}
\textbf{Chittaranjan Hota}
Chittaranjan Hota is a Senior Professor of Computer Science at BITS Pilani, Hyderabad, with 35 years of experience in teaching, research, and academic leadership. He holds a PhD from BITS Pilani and has over 150 publications, supervising 18+ PhD students. His research spans Big Data Analytics, Cybersecurity, and Machine Learning, with a focus on secure Bio-Cyber-Physical Systems under India's National Mission on Cyber-Physical Systems. He has held key administrative roles, including serving as Associate Dean of Admissions and as the founding Head of the Computer Science Department at BITS Hyderabad. He has led multiple government and industry-funded projects on network security, code tamper-proofing, smart healthcare, and cognitive IoT. He has held visiting positions at leading global universities and has received prestigious awards, including the Australian Vice-Chancellors Award and two Erasmus Mundus Fellowships. He is a Senior Member of IEEE.
\endbio

\end{document}